\documentclass[manuscript,screen]{acmart}
\usepackage{subcaption}
\usepackage{multirow}

\AtBeginDocument{%
  }

\setcopyright{none}
\begin{document}

\title{Chinese Short-Form Creative Content Generation via Explanation-Oriented Multi-Objective Optimization}

\author{Shanlin Zhou}
\authornote{Both authors contributed equally to this research.}
\email{zhoushanlin@tongji.edu.cn}
\orcid{0000-0003-1016-1281}
\author{Xinpeng Wang}
\authornotemark[1]
\email{wangxinpeng@tongji.edu.cn}
\orcid{0000-0003-1857-0346}
\affiliation{%
  \institution{Tongji University}
  \city{Jiading Qu}
  \state{Shanghai Shi}
  \country{China}
}

\author{Yongtao Hao}
\authornote{Corresponding author.}
\orcid{0000-0001-8738-7866}
\affiliation{%
  \institution{Tongji University}
  \city{Jiading Qu}
  \state{Shanghai Shi}
  \country{China}
}
\email{haoyt@tongji.edu.cn}

\author{Jianxun Lian}
\email{Jianxun.Lian@microsoft.com}
\orcid{0000-0003-3108-5601}
\affiliation{%
  \institution{Microsoft Research Asia}
  \city{Haidian Qu}
  \state{Beijing Shi}
  \country{China}
}

\author{Zhenghao Liu}
\orcid{0000-0003-0083-3224}
\affiliation{%
 \institution{Northeastern University}
 \city{Shenyang}
 \country{China}}
\email{liuzhenghao@mail.neu.edu.cn}

\author{Laks V.S. Lakshmanan}
\orcid{0000-0002-9775-4241}
\email{laks@cs.ubc.ca}
\affiliation{%
  \institution{University of British Columbia}
  \city{Vancouver}
  \state{BC}
  \country{Canada}
}

\author{Xiaoyuan Yi}
\orcid{0000-0003-2710-1613}
\authornotemark[2]
\email{xiaoyuanyi@microsoft.com}
\affiliation{%
  \institution{Microsoft Research Asia}
   \city{Haidian Qu}
  \state{Beijing Shi}
  \country{China}
}
\renewcommand{\shortauthors}{Zhou et al.}

\begin{abstract}
Chinese demonstrates high semantic compactness and rich metaphorical expressiveness, enabling limited text to convey dense meanings while increasing the difficulty of generation and verification, particularly in short-form creative natural language generation (CNLG). In the real world, users often require personalized, fine-grained creative constraints, making reliable verification critical to guiding optimization. According to Brunswik's Lens Model from psychology, constraints' achievement can be inferred from sufficient observable cues. Existing studies are mainly outcome-oriented, implicitly assuming that the outcome itself provides adequate cues for verification. However, this assumption breaks down in Chinese short-form CNLG (e.g., naming or advertising) with diverse personalized constraints, where extremely brief outcomes inherently offer limited information. Explanations can naturally serve as extra cues. Nevertheless, under complex constraints, LLMs' explanations may suffer from hallucination, incompleteness, or ambiguity. To address these, we novelly formalize the Chinese short-form CNLG task as a heterogeneous multi-objective optimization (HMO) issue that needs to jointly optimize multiple personalized constraints and explanation reliability. We further propose MAGIC-HMO, a training-free multi-agent framework that optimizes these objectives through iterative generation and verification under an explanation-oriented multi-objective strategy. Experiments on \emph{Chinese Baby Naming}, a challenging benchmark, demonstrate that MAGIC-HMO significantly outperforms six strong baselines across various LLM backbones. Relevant data and codes are available at \url{https://github.com/foolfun/MAGIC_HMO}.
\end{abstract}




\maketitle
\section{Introduction}
\begin{figure}[htpb!] 
\centering 
\includegraphics[scale=0.35]{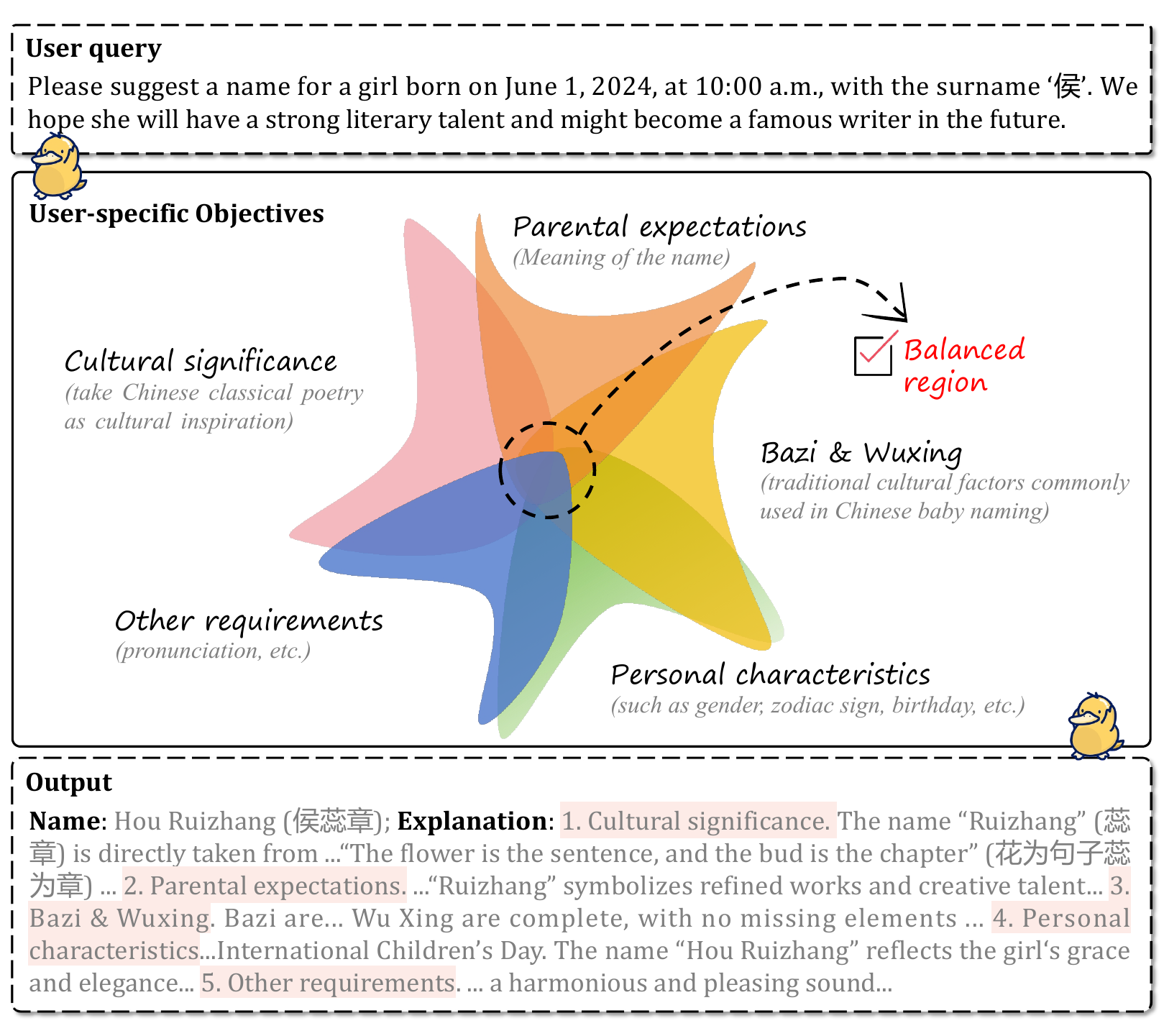} 
\caption{Example of Chinese baby naming. Different colors indicate diverse objectives.}
\Description{The figure depicts the process of generating a Chinese name with user-specific objectives. A user query specifies basic requirements such as surname, birth time, and expectations for future literary talent. Multiple objectives are considered, including cultural significance derived from classical poetry, parental expectations reflecting the intended meaning of the name, traditional factors such as Bazi and Wuxing, personal characteristics like gender and zodiac, and additional requirements such as pronunciation. These factors are visualized as overlapping regions, with a highlighted balanced region indicating the desired trade-off among competing objectives. The lower section shows an example output from a name generation system, including the generated name and a structured explanation covering cultural references, alignment with parental expectations, traditional compatibility, personal traits, and other linguistic considerations.}
\label{naming_example}
\end{figure}

As a specialized form of controllable generation~\cite{qian-etal-2022-controllable}, creative natural language generation (CNLG) aims to produce novel and aesthetically appealing text under user-specified constraints. Recently, large language models (LLMs)~\cite{annepaka2025large,li2025knowledge} have achieved remarkable progress in CNLG~\cite{10.1145/3592791,10.1145/3718331,10.1145/3708535,10.1145/3593805,10.1145/3637551}, enabling a wide range of applications such as poetry~\cite{chatzikyriakidis2025poetry,10.1145/3718331,10.1145/3593805}, summarization~\cite{laban2024summary}, and slogan generation~\cite{ahmad2024enhancing,10.1145/3637551}. 
In real-world applications, users are increasingly concerned with personalized objectives rather than general ones (e.g., novelty and fluency)~\cite{laban2024summary,chatzikyriakidis2025poetry}. For example, in Chinese naming scenarios, people typically consider factors such as parental expectations, Bazi \& Wuxing compatibility, personal traits, and cultural significance, as illustrated in Figure~\ref{naming_example}. While existing work has made progress in controlling general objectives, relatively limited attention has been paid to such personalized objectives.
Moreover, in short-form CNLG tasks, these personalized objectives are often not directly observable from the generated outcomes. For instance, given a name such as ``Hou Ruizhang'' in Figure~\ref{naming_example}, it is difficult to infer whether user-specific constraints (e.g., parental expectations) are satisfied based only on the name.  As a result, it is hard to determine whether a model genuinely understands based on a short outcome and satisfies user-specific requirements or only generates superficially plausible outcomes. This makes reliable verification and achievement of personalized objectives challenging in short-form CNLG.

This issue becomes more pronounced in Chinese short-form CNLG tasks (e.g., naming or advertising)~\cite{10.1145/3637551}. Due to the high semantic density and rich metaphorical expressiveness of Chinese, an extremely brief outcome can encode multiple user intents, yet provide only limited textual evidence for verifying objective satisfaction. From the perspective of Brunswik’s Lens Model~\cite{wolf2005brunswik}, individuals infer latent criteria through observable cues. This suggests the need to introduce additional cues for objective verification in Chinese short-form CNLG. 
Explanations naturally serve as external cues, helping reveal the latent semantics behind generated outcomes and facilitating objective verification. As illustrated in Figure~\ref{naming_example}, by adding structured explanations corresponding to user-specified objectives to the generated name  ``Hou Ruizhang'', it becomes possible to more clearly interpret whether these underlying constraints are satisfied. 
However, LLM-generated explanations often suffer from hallucination, incompleteness, and ambiguity, especially under complex or conflicting constraints~\cite{zhao2024explainability}.
In conclusion, Chinese short-form CNLG faces the following two unique challenges:
\begin{itemize}
    \item \textit{Challenge 1: Personalized requirements satisfaction}. The limited textual outcome makes it difficult for models to fully capture and satisfy multiple user-specific constraints, often leading to partial or imprecise alignment with user requirements~\cite {kumar2021controlled,pham-etal-2024-suri}. 
    \item \textit{Challenge 2: Reliable explanations generation}. While explanations can improve interpretability and help constraint verification, LLMs frequently produce them with hallucination, incompleteness, or ambiguity, thereby undermining their reliability~\cite{zhao2024explainability}.
\end{itemize}

\begin{figure}[htpb!] 
\centering 
\includegraphics[scale=0.5]{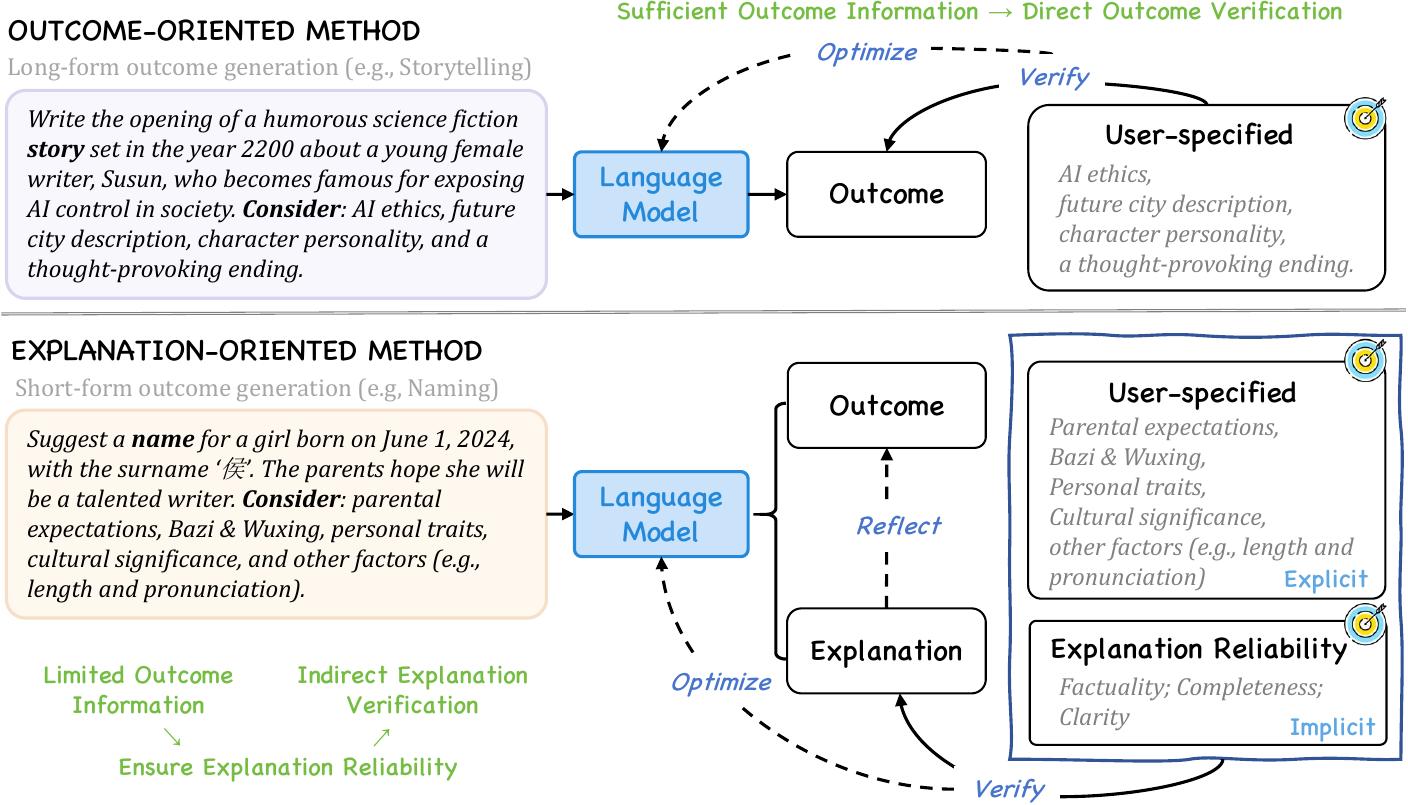} 
\caption{Comparison between outcome-oriented (top) and explanation-oriented (bottom) methods for multi-objective creative text generation. The top section highlights approaches that focus primarily on optimizing final outputs without explicitly modeling the underlying reasoning process, while the bottom section illustrates methods that incorporate structured explanations to improve transparency, controllability, and alignment with multiple user-specified objectives.}
\Description{The figure compares two methods for multi-objective creative text generation. The top panel illustrates an outcome-oriented method, where the model directly generates a creative outcome based on user input. The generated outcome is iteratively optimized and verified against user-specified criteria, assuming that the outcome has sufficient information to verify the objective satisfaction directly. The bottom panel presents an explanation-oriented method designed for scenarios with a limited length of outcome. The model produces both the outcome and a structured explanation that reflects multiple objectives decided by the user. Instead of relying on the outcome-oriented optimization, this method indirectly verifies the objectives and improves the outcome by constructing and optimizing a set of accurate, complete, and clear explanations.}
\label{fig_0}
\end{figure}
Existing approaches on multi-aspect controllable text generation, including  decoding~\cite{ding-etal-2023-maclasa}, fine-tuning~\cite{pham2024suri}, reinforcement learning~\cite{zhou2024beyond}, and agent-based methods (e.g., LLM-D~\cite{lu2024llm}). These methods primarily adopt an outcome-oriented optimization paradigm, as shown in Figure~\ref{fig_0}, which evaluate and optimize user-specified objectives only based on generated outcomes. Outcome-oriented optimization methods implicitly assume that the outcomes themselves provide sufficient observable cues for verifying objective satisfaction. While this assumption may hold for long-form CNLG, where richer textual content offers more cues for verification, it breaks down in Chinese short-form scenarios with limited evidence. To address this limitation, we propose shifting from outcome-oriented optimization to an explanation-oriented paradigm (Figure~\ref{fig_0}). Instead of relying only on outcomes, our provided paradigm treats explanations as intermediate and interpretable signals for verification. Specifically, we first ask the model to generate structured explanations that correspond to user-specified objectives, and then evaluate their reliability, and leverage the verified explanations as indirect cues to evaluate whether these objectives are satisfied. This enables more effective and interpretable multi-objective optimization when sufficient cues cannot be directly inferred from the generated outcomes.

We formalize Chinese short-form CNLG as a \textit{Heterogeneous Multi-objective Optimization (HMO)} problem, including two sets of heterogeneous objectives. \emph{Explicit User-specified Objectives (EUOs)} refer to user-specified creative requirements in task description. 
\emph{Implicit Reliability Objectives (IROs)} characterize essential reliability of explanations, including factuality, completeness, and clarity. 
These correspond to the two challenges mentioned above. 
To address these, we further propose \textbf{MAGIC-HMO}, a training-free \underline{M}ulti-\underline{A}gent collaborative framework for \underline{G}enerating short-form \underline{I}nnovative \underline{C}ontent by explain-oriented verification and optimization strategy for HMO. MAGIC-HMO aims to balance EUOs, IROs, and their trade-offs. This framework consists of three cooperative agents (Figure~\ref{fig_02}), including a multi-objective manager (MOM), a multi-objective generator (MOG), and a multi-objective evaluator (MOE). MOM performs task analysis and information processing around personalized requirements. MOG generates outcomes and explanations by integrating this information. MOE evaluates both intermediate reasoning and final outputs.
For \textit{Challenge~1}, we address it by enhancing the input to MOG through MOM and leveraging MOE to evaluate the satisfaction of EUOs in outputs.
For \textit{Challenge~2}, we design an evaluation mechanism centered on explanation quality and unreliable explanations trigger corrective feedback to refine IROs. 
In summary, MAGIC-HMO realizes HMO through two stages: (i) information enhancement, and (ii) explanation-oriented dynamic iterative optimization that constructs reliable explanations to understand and guide optimization of creative outcomes.

To demonstrate the effectiveness of our framework in a realistic and challenging scenario, we treat \emph{Chinese Baby Naming (CBN)} as a representative task of Chinese short-form CNLG. As illustrated in Figure~\ref{naming_example}, a common Chinese name may contain multiple meanings simultaneously, such as traditional cultural significance, parents’ expectations, Bazi \& Wuxing\footnote{Bazi \& Wuxing are traditional cultural factors commonly used in CBN. Bazi analyzes the birth date and time, while Wuxing represents the five elements (wood, fire, earth, metal, and water). Together, they provide insight into fate and fortune based on traditional Chinese numerology.}, and personal characteristics. Since the need to compress implicit meanings and aesthetic qualities into an extremely short name, high-quality explanations become important to understand these. Therefore, CBN is a challenging and ideal short-form CNLG task for studying explanation-oriented HMO. We further develop a benchmark named \emph{CBNames}. Given cultural meaning (\textit{e.g.}, aesthetics and poeticness) is often considered in names, but LLMs often produce hallucinations, we further collect an \emph{ancient Chinese poetry dataset (CPoetry)}. Moreover, some new metrics are designed for reasonable and comprehensive evaluation. Experiments on various backbones reveal the excellence of MAGIC-HMO over six strong baselines in short-term multi-objective CNLG. 

In summary, our contributions are as follows: 
\begin{itemize}
\item To the best of our knowledge, we are the first to identify challenges in Chinese short-form CNLG and novelly formalize it as heterogeneous multi-objective optimization (HMO).
\item  For HMO, we propose MAGIC-HMO, a training-free multi-agent collaborative framework for reliable and interpretable Chinese short-form creative generation by an explanation-oriented multi-objective optimization strategy.
\item We construct two new datasets: \textit{CBNames} for the Chinese Baby Naming (CBN) task and \textit{CPoetry} for classical poetry retrieval. Extensive experiments on six backbones indicate that MAGIC-HMO outperforms existing methods in creative quality.
\end{itemize}

\section{Related Work}
\paragraph{Creative Natural Language Generation}
Creative Natural Language Generation (CNLG) emphasizes the novelty, distinctiveness, and aesthetic quality of generated text, enabling a wide range of downstream applications such as poetry generation~\cite{zhipeng2019jiuge,yu2024charpoet,chatzikyriakidis2025poetry,10.1145/3718331,10.1145/3593805}, story writing~\cite{venkatraman-etal-2025-collabstory,yang2024seed,10.1145/3708886}, summarization~\cite{laban2024summary,zhang2025comprehensive}, and slogan generation~\cite{ahmad2024enhancing,10.1145/3637551}. Traditional studies~\cite{zhipeng2019jiuge,yu2024charpoet,venkatraman-etal-2025-collabstory} in this area primarily focus on optimizing intrinsic properties of generated text, including novelty, originality, fluency, and diversity. With the rapid advancement of large language models (LLMs), CNLG systems have achieved substantial improvements in both generation quality and scalability, leading to increasingly widespread adoption in real-world applications such as marketing, content creation, and human-computer interaction~\cite{annepaka2025large,10976701}.
However, as CNLG systems are increasingly deployed in practical and user-centric scenarios, user instructions have become more complex, often involving fine-grained, personalized, and multi-dimensional constraints~\cite{10.1145/3708886,10.1145/3593805}. These constraints pose new challenges for creative generation. The high semantic density and rich expressive flexibility of the Chinese language in a short-form text further amplify this difficulty, making it more challenging for models. The limited textual space intensifies competition among constraints and raises the risk of omission or conflict. As a result, balancing multiple personalized creativity objectives becomes more challenging for models (\emph{challenge 1}).

\paragraph{Multi-Aspect Controllable Text Generation} 
Multi-Aspect Controllable Text Generation (MCTG) aims to generate text that satisfies multiple constraints (e.g., sentiment, topic, style). Traditional methods often focus on single-aspect control, but extending to multi-aspect scenarios introduces challenges like attribute interference and degeneration~\cite{khalifa2021a,gu2022distributional,huang2023extensible,ding-etal-2023-maclasa,gu-etal-2023-controllable,cao2024tara}. Recent studies~\cite{khalifa2021a,gu2022distributional,huang2023extensible,ding-etal-2023-maclasa,gu-etal-2023-controllable,cao2024tara,pham2024suri,wen2024benchmarking,he2024complex} have begun to explore more diverse objectives for control. Typical methods include MacLaSa~\cite{ding-etal-2023-maclasa}, Mix\&Match~\cite{mireshghallah-etal-2022-mix}, COLD~\cite{qin2022cold} and  MUCOCO~\cite{kumar2021controlled} for decoding-based methods, as well as fine-tuning~\cite{pham2024suri, he2024complex}, reinforcement learning~\cite{zhou2024beyond}, and agent-based methods like LLM-D~\cite{lu2024llm}. However, these methods typically focus on optimizing outcomes by outcome-oriented evaluation. This results in a gap in understanding the hidden meanings that are crucial for short-form CNLG tasks. Due to extremely brief outcomes in short-form CNLG, reliable explanations are necessary to verify whether all user-specified objectives are truly met (\emph{challenge 2}). Thus, we propose an explanation-oriented heterogeneous multi-objective optimization framework, which not only optimizes the generated content but also ensures that the reasoning process behind the outcomes is interpretable and verifiable.

\paragraph{LLM-Based Multi-Agent.}
Given the strong reasoning and planning capabilities of large language models (LLMs), LLM-based agents have seen rapid development.
Compared to single-agent systems that solve problems through tool use and interactions with external environments, multi-agent systems emphasize decision-making through role-playing and communication among multiple agents~\cite{du2024improving,park2023generative,xu2024exploringlargelanguagemodels}. 
However, the use of multi-agent systems has primarily focused on solving complex tasks~\cite{9930851,9712868,10738426}. To the best of our knowledge, no prior work has explored leveraging multi-agent collaboration to explicitly address multi-objective tasks, which is a gap our work seeks to fill.
\section{Methodology} 
\subsection{Overview}
In this section, we formally define the heterogeneous multi-objective optimization (HMO) problem in Chinese short-form creative natural language generation (CNLG) and the proposed explanation-oriented method.

Following the Lens Model in cognitive psychology, humans infer the state of latent objectives based on observable cues $C$. In Chinese short-form CNLG, given a user query $U$ containing $m$ personalized creative constraints, the true satisfaction of user requirements is hard to directly observe through a short outcome $y$ as the only cue. We introduce the corresponding explanations $e=\{e_1, e_2, \dots, e_m\}$ as extra cues to help improve its quality and explainability. In Chinese short-form CNLG, it is crucial to construct reliable explanations to understand and guide the optimization of creative results. Therefore, we view Chinese short-form CNLG as HMO, which aims to optimize explicit user-specific objectives (EUOs), implicit explanation reliability objectives (IROs), and the trade-offs between them. As a result, HMO can be formulated as: 
\begin{equation}
\begin{aligned}
(y_t^*, e_t^*) =
\arg\max_{y,e} \;
\big(&F_{\text{EUO}}(y,e \mid U;\theta_t), &F_{\text{IRO}}(y,e \mid U;\theta_t)\big)
\end{aligned}
\end{equation}
where $\theta$ is a dynamic optimization parameter. $F_{\text{EUO}}(y,e)$ measures the satisfaction of the explicit user-specific objectives, while $F_{\text{IRO}}(y,e)$ evaluates the reliability of the explanations $e$. 

\begin{figure}[htpb!] 
\centering 
\includegraphics[width=\linewidth]{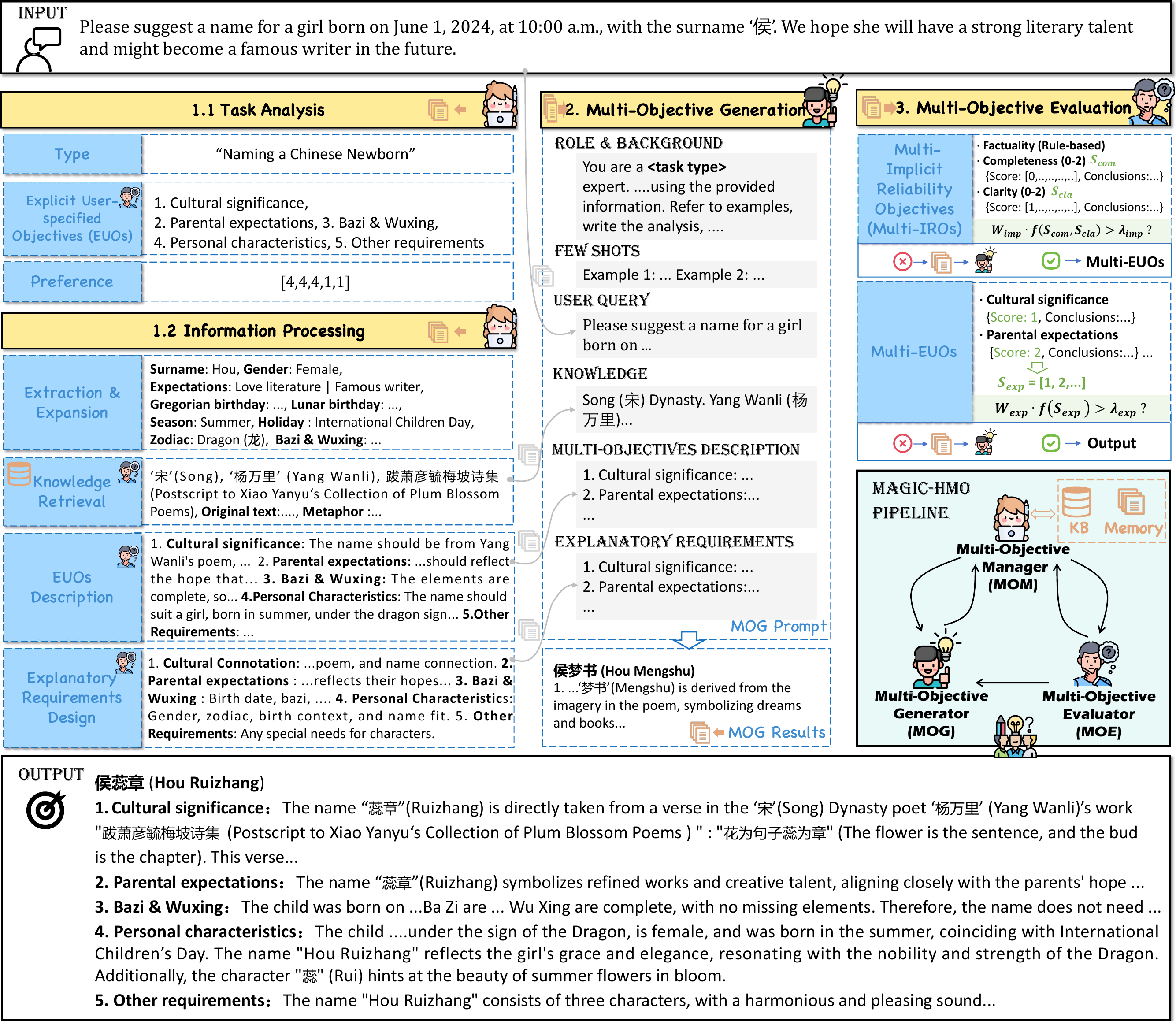} 
\caption{Overview of MAGIC-HMO.  Steps 1.1 and 1.2 constitute the multi-objective information preparation process, which is primarily handled by MOM and MOE. The dynamic iterative objective optimization process includes Steps 2 and 3: Step 2 is managed by MOG, while Step 3 reflects MOE’s role in evaluating the generation outcomes. The green block at the bottom right illustrates the complete pipeline of MAGIC-HMO.}
\Description{The figure presents a pipeline for explain-oriented multi-objective generation and evaluation. The process begins with an input query describing the naming task and user expectations. In the task analysis stage, the task type and explicit user objectives are identified, including cultural significance, parental expectations, traditional compatibility, personal characteristics, and other constraints, along with their relative preferences. During information processing, relevant attributes such as surname, gender, birth details, seasonal context, and external knowledge sources are extracted and expanded. These are used to construct structured descriptions of objectives and corresponding explanatory requirements. In the multi-objective generation stage, a generator produces candidate outputs guided by role definitions, few-shot examples, user queries, and retrieved knowledge. The outputs include both generated names and detailed explanations aligned with the defined objectives. In the evaluation stage, multiple implicit reliability objectives such as factuality, completeness, and clarity are assessed, alongside explicit objective satisfaction scores. A selection mechanism determines whether outputs meet predefined thresholds. The pipeline is coordinated by a multi-objective manager that iteratively refines generation and evaluation, supported by memory and knowledge bases. The final output includes a generated name addressing all objectives with a set of clear and structured explanations.}
\label{fig_02}
\end{figure}
In our work, we design a training-free multi-agent architecture, MAGIC-HMO, to address the HMO problem via an explanation-oriented strategy. Our pipeline is executed by three intelligent agents, including Multi-Objective Manager (MOM) $A_\text{M}$, Multi-Objective Generator (MOG)  $A_\text{G}$, and Multi-Objective Evaluator (MOE) $A_\text{E}$ (green block of Figure~\ref{fig_02}). The pipeline consists of two stages. \textbf{Stage 1: Information enhancement} (steps 1.1–1.2 in Figure~\ref{fig_02}) is performed by MOM $A_\text{M}$ and MOE $A_\text{E}$ to prepare heterogeneous information for multi-objective generation. \textbf{Stage 2: Explanation-oriented dynamic iterative optimization} (steps 2–3 in Figure~\ref{fig_02}) iteratively refines the output by optimizing explanations. Specifically, MOG $A_\text{G}$ first generates a draft using information from MOM $A_\text{M}$, which is then adaptively refined based on MOE $A_\text{E}$'s feedback on EUOs and IROs. Next, we describe each stage in detail.
\subsection{Information Enhancement}
This stage performs task analysis and key information expansion to better understand user-specific requirements and enhance the model's ability to interpret implicit meaning, helping to produce higher-quality creative generation.
\paragraph{Task Analysis} 
The process of task analysis (see 1.1 in Figure~\ref{fig_02}) includes identifying the task type (e.g., Chinese baby naming), getting EUOs, and estimating user preference scores for EUOs. Specifically, when EUOs are not provided by the user, MOM $A_\text{M}$ can automatically parse appropriate EUOs. However, when auto-parsing objectives, the model may parse similar ones (e.g., "Harmonious Pronunciation", "Easy to Pronounce and Remember", etc.). Therefore, we introduce MOE $A_\text{E}$ to help this decision-making process, ensuring the independence and appropriateness of each objective. We estimate user preference by $A_\text{M}(\text{EUOs}, U)$ and normalize it into a weight vector $W_{\text{exp}}$. This helps better meet different personalized requirements. Later, this weight $W_\text{exp}$ will work in the next stage.
\paragraph{Information Expansion}
The information expansion process (1.2 in Figure~\ref{fig_02}) covers extracting and expanding information, retrieving related knowledge, and refining the EUOs' description and explanatory requirements.
Firstly, extracting key content $I_{\text{ki}} \gets A_\text{M}(U)$ (e.g., surname and date of birth in the naming task) prepares for retrieval. If $I_{\text{ki}}$ is insufficient, it can infer additional details (e.g., deducing the birth season based on the date of birth) to enrich it.
To mitigate domain hallucination in LLMs, we introduce a retrieval step. The multi-aspect requirements in user input $U$ make the retrieval process inherently multi-objective. Specifically, MOM first uses the key information $I_\text{ki}$ to retrieve a set of preliminary candidates $R'$ from the knowledge base $D$. Then, it reconstructs a style-aligned query $q$ based on $U$ and retrieve the top-$k$ candidates $R = \{r_{1}, r_{2}, \ldots, r_{k}\}$ from $R'$, using an embedding-based retrieval strategy~\footnote{https://huggingface.co/lier007/xiaobu-embedding-v2}. Style-reshaped query can help to increase the efficiency of retrieval. To find the most relevant knowledge $I_\text{rk}$, MOE participates in evaluation and reranking. If no candidate is satisfactory, MOE revises the query $q$ while keeps its style. The revised query is fed back to MOM, which removes previously retrieved candidates $H$ from the knowledge base $D$ and performs a new retrieval over the remaining $D$. This iterates until the optimal $I_\text{rk}$ is found.
To help MOG better understand and generate output align with EUOs and IROs, MOM $A_\text{M}$  constructs detailed descriptions $I_{\text{desc}}=\{c_{1},c_{2}, \ldots, c_{m}\}$ and explanatory requirements $I_{\text{reqs}}=\{q_{1},q_{2}, \ldots, q_{m}\}$. Descriptions of EUOs provide more detailed semantic information, and fine-grained requirements of IROs help MOG generate more accurate, complete, and clear content. MOE $A_\text{E}$ participates in this process by evaluating and refining these contents through feedback and iteration. These steps are as follows:
\begin{equation}
     (I_{\text{desc}}, I_{\text{reqs}}) \gets A_\text{M}(U, \text{EUOs}, I_{\text{ki}}, I_{\text{rk}},b_{\text{detail}})
\end{equation}
\begin{equation}
    b_{\text{detail}} \gets A_{E}(I_{\text{desc}}, I_{\text{reqs}})
\end{equation}
where $b_{detail}$ is the evaluation feedback. Moreover, they are also important criteria in the dynamic optimization stage.
Finally, we obtain the expanded information set $I = \{I_\text{type}, I_\text{pref}, I_\text{rk}, I_\text{desc}, I_\text{reqs}\}$ and store it in MOM. 

\subsection{Explanation-oriented Dynamic Iterative Optimization}
MOG utilizes user query $U$ and expanded information $I$ to generate the initial output $(y,e)$. Then, MOE evaluates the alignment of $(y,e)$ with EUOs and IROs separately according to $I_{\text{desc}}$ and $I_{\text{reqs}}$, and feeds the evaluation results back to MOG.

IROs are evaluated first, as reliable explanations are a prerequisite for verifying EUOs. Specifically, we define three IROs, including accuracy, completeness, and clarity. Accuracy is measured using a rule-based function with reference to information $I$, which helps mitigate hallucinations. Given the subjectivity of completeness and clarity, MOE employs scoring mechanisms with $I_{\text{reqs}}$ to evaluate each explanation on these aspects. The scores are averaged, normalized, and aggregated to obtain the final implicit objective score $\theta_{\text{imp}}$:
\begin{equation}
     \theta_\text{imp} = W_\text{imp} \cdot S_\text{imp}
\end{equation}
\begin{equation}
    W_{\text{imp}} = [\frac{1}{n}, \frac{1}{n}, \dots, \frac{1}{n} ] \in \mathbb{R}^n
\end{equation}
where $W_{\text{imp}}$ is the weight vector,  $S_\text{imp}$ is the IROs score vector. This formulation converts the optimization of IROs into a single-objective problem.

Similarly, MOE evaluates the performance of explanations on EUOs using the descriptions $I_{\text{desc}}$ and combines the results into a score vector $S_\text{exp}$. Since the user input $U$ often implies preferences over different EUOs, the preference information $I_\text{pref}$ can be the optimization weight vector $W_\text{exp}$. The unified EUOs score is computed as:
\begin{equation}
     \theta_\text{exp} = W_\text{exp} \cdot S_\text{exp}
\end{equation}
Thresholds $\psi_\text{imp}$ and $\psi_\text{exp}$ are introduced to judge implicit and explicit objective scores, respectively, and are dynamically adjusted during iteration:
\begin{equation}
\psi_{\text{imp}} = 
\begin{cases}
\delta & j_{\text{imp}} < t_w \\
\frac{\delta}{\alpha \cdot \log(j_{\text{imp}} + t_\text{w})} & j_{\text{imp}} > t_\text{w}
\end{cases}
\end{equation}
\begin{equation}
\psi_{\text{exp}} = 
\begin{cases}
\delta & j_{\text{exp}} < t_w \\
\frac{\delta}{\alpha \cdot \log(j_{\text{exp}} + t_\text{w})} & j_{\text{exp}} > t_\text{w}
\end{cases}
\end{equation}
where $j_{\text{imp}}$ and $j_{\text{exp}}$ denote the evaluation round counters for implicit and explicit objectives, respectively, $t_w$ is the warmup round, $\delta$ is the initial threshold, and $\alpha$ is the decay factor. Thanks to this dynamic adjustment, the reliability and quality of output are improved while multiple explicit objectives are achieved in balance. 
\section{Experiments}
\subsection{Experimental Setup}
\subsubsection{Datasets}
To validate the effectiveness of our framework, we study \emph{Chinese Baby Naming (CBN)}, a representative Chinese short-term Creative Natural Language Generation (CNLG) task. CBN requires considering multiple user-specific objectives when generating a short name, making it particularly challenging. We construct a benchmark named \textbf{CNames} by simulating 500 human naming requests with a large language model (LLM) and manual verification. The dataset covers over 200 different Chinese surnames and reflects diverse user requirements, such as cultural meanings, parental expectations, the Bazi\&Wuxing, personal characteristics, and other special requirements, with user preferences annotated per objective.
Since classical poetry is a common source of culturally meaningful names and LLMs often make factual errors in this domain, we further compile \textbf{CPoetry}, a dataset of 176,450 classical Chinese poems with metadata including poet, dynasty, title, content, interpretations, and themes.
\subsubsection{Baseline Methods}
We compare our proposed framework, MAGIC-HMO, with the following methods. \textbf{Base} feeds the raw query with user constraints into the model. \textbf{CoT}~\cite{kojima2022large} appends the phrase “Let’s think step by step” to the prompt. \textbf{TDB}~\cite{yang2024largelanguagemodelsoptimizers} uses the zero-shot prompt “Take a deep breath and work on this problem step-by-step”. \textbf{Few-shot}~\cite{brown2020language} adds a few demonstration examples in the prompt. \textbf{Q2Kw}~\cite{jagerman2023query} retrieves initial results using the raw query, forms an expanded query by combining them with the original query, and incorporates the retrieved knowledge into the prompt. \textbf{LLM-D}~\cite{lu2024llm} employs multiple agents to conduct multi-stage discussions for generating creative responses.
\subsubsection{Evaluation and Metrics}
Recent research has shown the capability of LLM to emulate human judgment and effectively evaluate content~\cite{li2024llms,adlakha2024evaluating}. In this work, we evaluate quantitatively using both LLM and human evaluations. 

\textbf{In LLM evaluation}, we propose metrics to evaluate explicit user-specified objectives (EUOs), implicit reliability objectives (IROs), and their joint optimization, and to measure how well different methods perform in the short-term CNLG task.
Firstly, for EUOs, we design \emph{\underline{e}xplicit multi-objective \underline{c}ompleteness} (\textbf{EC}) and its standard deviation(\textbf{EC\_std}):
\begin{equation} 
\text{EC} = \frac{1}{N} \sum_{i=1}^{N} \frac{\sum_{j=1}^{m} w_{i,j} s_{i,j}}{\sum_{j=1}^{m} w_{i,j}}
\end{equation}
\begin{equation}
    \text{EC\_std} = \frac{1}{N} \sum_{i=1}^{N} \sqrt{ \frac{1}{m_{i}} \sum_{j=1}^{m_{i}} \left( s_{i,j} - \frac{1}{m_{i}} \sum_{j=1}^{m_{i}} s_{i,j} \right)^2 }
\end{equation}
where $s_{i,j}$ and $w_{i,j}$ denote the score and user weight for the $j$-th objective of $i$-th sample, with $m$ EUOs across $N$ samples. Higher EC indicates better performance in satisfying EUOs, while a lower EC\_std indicates better balance across EUOs.
For IROs, we adopt \textbf{ACC}, \textbf{CRC}, and \textbf{LC}, which measure hallucination degree, semantic quality (comprehensiveness, relevance, clarity), and logical consistency, respectively. Higher scores indicate better performance. We further define \emph{\underline{i}mplicit multi-objective \underline{c}ompleteness} (\textbf{IC}) as the average of ACC, CRC, and LC, and \textbf{IC\_std} as their standard deviation:
\begin{equation}
    \text{IC\_std} = \frac{1}{N} \sum_{i=1}^{N} \sqrt{ \frac{1}{3} \sum_{k \in \{ACC, CRC, LC\}} \left( s_{i,k} - \bar{s}_i \right)^2 }
\end{equation}
where $\bar{s}_i$ is the mean score of the $i$-th sample across ACC, CRC, and LC, with $s_{i,k}$ denoting the score for the $k$-th implicit objective. Higher IC indicates better IROs performance, while lower IC\_std reflects more balanced results.
Finally, from a holistic perspective, we define \emph{\underline{c}omprehensive multi-objective \underline{c}ompleteness} (\textbf{CC}) as the average of EC and IC, \textbf{CC\_std} as their corresponding standard deviations. 
\begin{equation}
    \text{CC\_std} = \frac{1}{N} \sum_{i=1}^{N} \sigma_i
\end{equation}
\begin{equation}
    \sigma_i = \sqrt{\frac{1}{2} \left( (\text{EC}_i - \bar{x}_i)^2 + (\text{IC}_i - \bar{x}_i)^2 \right)}
\end{equation}
where $\sigma_i$ denotes the standard deviation of EC and IC for sample $i$, and $\bar{x}_i$ is their mean value. Moreover, we use \underline{div}ersity (\textbf{DIV}) as the uniqueness of results across different methods for each sample: if a result also appears in other methods, it receives a score of 0; otherwise, 1. Higher DIV indicates better diversity. 
In short, EC and EC\_std reflect the performance and balance of multi-EUO optimization. ACC, CRC, and LC evaluate explanation quality, while IC and IC\_std measure the performance and balance of multi-IRO optimization. CC and CC\_std capture the overall completeness and balance across both EUOs and IROs, and DIV reflects the diversity of results across methods.

\textbf{In human evaluation}, we randomly sample 50 test cases, each evaluated by three annotators on a 0–3 scale. EUOs are scored based on completeness, while IROs are evaluated in CRC and LC. Annotators also select the best response among the compared methods for each case. For consistency, all responses evaluated by annotators are generated based on Qwen as the backbone model.
\subsubsection{Implementation Details}
To ensure consistency, we specify five objectives for CBN, including cultural meaning, parental expectations, Bazi\&Wuxing, personal characteristics, and other special requirements. We evaluate seven backbone LLMs: Qwen (long)~\cite{bai2023qwentechnicalreport}, GLM-4~\cite{team2024gemini}, DeepSeek (V3)~\cite{deepseekai2025deepseekv3}, Mistral (small-latest)~\cite{jiang2023mistral7b}, Gemini (1.5-flash)~\cite{team2024gemini}, and GPT4o~\cite{openai2024gpt4technicalreport}. Qwen, GLM-4, and DeepSeek are stronger in Chinese, while Mistral, Gemini, and GPT-4o perform better in English. DeepSeek uses the recommended temperature of 1.5 to encourage creativity. We use Kimi~\cite{team2025kimi} as the LLM evaluator. ACC is computed via LLM extraction combined with rule-based verification. Other metrics are rated on a 0–3 scale (0 = invalid, 3 = excellent), then normalized to 0–100.
\subsection{Method Comparison}
\begin{table}[htpb!]
\caption{LLM Evaluation Results. Bold ones indicate the best-performing method for each metric.}
\resizebox{0.7\linewidth}{!}{
\begin{tabular}{clcccc|ccc}
\toprule

\multirow{2}{*}{Backbone}
& \multirow{2}{*}{Method}
& \multicolumn{2}{c}{\textbf{Explicit}}
& \multicolumn{2}{c}{\textbf{Implicit}}
& \multicolumn{3}{|c}{\textbf{Comprehensive}} \\
\cmidrule(r){3-4} \cmidrule(r){5-6} \cmidrule(r){7-9}
&  & EC ↑ & EC\_std ↓ & IC ↑ & IC\_std ↓ & CC ↑ & CC\_std ↓ & DIV ↑ \\

\midrule
\multicolumn{1}{c}{\multirow{7}{*}{Qwen}}
& Base	    & 85.03  & 18.75  & 76.29  & 11.28  & 80.66 & 8.52  & 44.20 \\
& CoT	    & 76.98  & 23.99  & 70.75  & 24.08  & 73.86 & 8.33  & 91.00 \\
& TDB	    & 87.34  & 15.52  & 82.85  & 11.45  & 85.10 & 6.61  & 44.80 \\
& Few-shot  & 94.07  & 9.74   & 76.43  & 16.59  & 85.25 & 12.70 & 60.40 \\
& Q2Kw	    & 83.31  & 19.22  & 80.05  & 11.93  & 81.68 & 6.59  & 84.60 \\
& LLM-D	    & 85.51  & 17.65  & 80.75  & 14.67  & 83.13 & 7.20  & 82.80 \\
& \textbf{MAGIC-HMO}	& \textbf{96.72} & \textbf{5.95} & \textbf{92.70} & \textbf{8.28}  & \textbf{94.71} & \textbf{4.64} & \textbf{99.80}\\

\midrule
\multicolumn{1}{c}{\multirow{7}{*}{GLM4}}
& Base      & 88.37  & 14.79  & 79.44  & 10.15  & 83.90  & 7.96   & 56.60 \\
& CoT       & 80.25  & 21.19  & 73.39  & 20.38  & 76.82  & 8.30   & 93.80 \\
& TDB       & 88.12  & 15.00  & 83.25  & 12.39  & 85.68  & 6.30   & 56.20 \\
& Few-shot  & 94.10  & 10.51  & 79.49  & 14.75  & 86.79  & 11.00  & 75.80 \\
& Q2Kw      & 85.95  & 18.39  & 80.40  & 11.19  & 83.18  & 6.93   & 91.40 \\
& LLM-D     & 91.21  & 12.55  & 86.21  & 11.48  & 88.71  & 6.03   & 96.20 \\
& \textbf{MAGIC-HMO}  & \textbf{97.83}  & \textbf{4.37}  & \textbf{92.94}  & \textbf{8.31}  & \textbf{95.38} & \textbf{4.45} & \textbf{98.80} \\

\midrule
\multicolumn{1}{c}{\multirow{7}{*}{DeepSeek}}
& Base      & 93.53  & 9.84   & 85.29  & 13.66  & 89.41  & 7.19   & 69.20 \\
& CoT       & 93.46  & 9.38   & 84.74  & 14.89  & 89.10  & 7.91   & 73.60 \\
& TDB       & 93.40  & 10.01  & 84.91  & 14.00  & 89.15  & 7.50   & 72.20 \\
& Few-shot  & 98.02  & 4.50   & 84.25  & 19.03  & 91.14  & 9.85   & 76.80 \\
& Q2Kw      & 93.17  & 10.28  & 86.93  & 11.43  & 90.05  & 6.54   & 96.00 \\
& LLM-D     & 93.90  & 8.81   & 81.21  & 14.59  & 87.56  & 10.31  & 96.60 \\
& \textbf{MAGIC-HMO}  & \textbf{98.93}  & \textbf{3.45} & \textbf{95.22}  & \textbf{5.96}   & \textbf{97.08} 	& \textbf{3.00}    & \textbf{99.60} \\

\midrule
\multicolumn{1}{c}{\multirow{7}{*}{Mistral}}
& Base      & 82.10  & 19.43  & 68.11  & 11.82  & 75.11  & 11.86  & 88.40 \\
& CoT       & 82.46  & 18.68  & 72.40  & 11.64  & 77.43  & 9.77   & 86.60 \\
& TDB       & 81.09  & 19.52  & 71.55  & 11.94  & 76.32  & 10.02  & 86.80 \\
& Few-shot  & 93.97  & 9.43   & 67.17  & 22.54  & 80.57  & 19.23  & 94.00 \\
& Q2Kw      & 79.82  & 21.85  & 70.81  & 15.11  & 75.31  & 9.56   & 91.60 \\
& LLM-D     & 84.19  & 17.56  & 75.88  & 12.43  & 80.03  & 8.95   & 97.00 \\
& \textbf{MAGIC-HMO}  & \textbf{94.94}  & \textbf{7.28} & \textbf{91.71}  & \textbf{9.56}   & \textbf{93.32} 	& \textbf{5.53}    & \textbf{98.80} \\

\midrule
\multicolumn{1}{c}{\multirow{7}{*}{Gemini}}
& Base      & 84.56  & 21.14  & 74.10  & 12.31  & 79.33  & 8.87   & 69.60 \\
& CoT       & 80.17  & 23.87  & 72.28  & 18.61  & 76.22  & 8.66   & 87.40 \\
& TDB       & 81.50  & 23.16  & 73.11  & 17.05  & 77.31  & 8.49   & 81.20 \\
& Few-shot  & 93.66  & 12.51  & 73.77  & 15.88  & 83.71  & 14.53  & 76.40 \\
& Q2Kw      & 82.81  & 22.03  & 77.41  & 14.41  & 80.11  & 7.20   & 76.60 \\
& LLM-D     & 82.15  & 21.93  & 76.84  & 15.75  & 79.49  & 7.61   & 87.60 \\
& \textbf{MAGIC-HMO}  & \textbf{97.51}  & \textbf{5.15} & \textbf{92.72}  & \textbf{8.76}   & \textbf{95.12} 	& \textbf{4.57}    & \textbf{99.20} \\

\midrule
\multicolumn{1}{c}{\multirow{7}{*}{GPT4o}}
& Base      & 86.08  & 17.34  & 79.29  & 13.87   & 82.69  & 7.15  & 94.80 \\
& CoT       & 78.10  & 23.31  & 71.90  & 19.90   & 75.00  & 8.08  & 96.80 \\
& TDB       & 80.07  & 22.60  & 72.81  & 18.45   & 76.44  & 8.26  & 96.20 \\
& Few-shot  & 95.34  & 7.95   & 77.68  & 22.88   & 86.51  & 12.80 & 96.00 \\
& Q2Kw      & 83.00  & 22.24  & 75.07  & 14.70   & 79.04  & 8.15  & 95.80 \\
& LLM-D     & 82.93  & 20.62  & 77.01  & 14.50   & 79.97  & 7.95  & 96.00 \\
& \textbf{MAGIC-HMO}  & \textbf{99.15}  & \textbf{2.32} & \textbf{96.22}  & \textbf{5.18}   & \textbf{97.69} 	& \textbf{2.55}    & \textbf{99.00} \\

\bottomrule
\end{tabular}}
\label{tab_com}
\end{table}
\begin{figure}[htpb!]
\centering
\begin{subfigure}{0.8\linewidth}
    \centering
    \includegraphics[width=\linewidth]{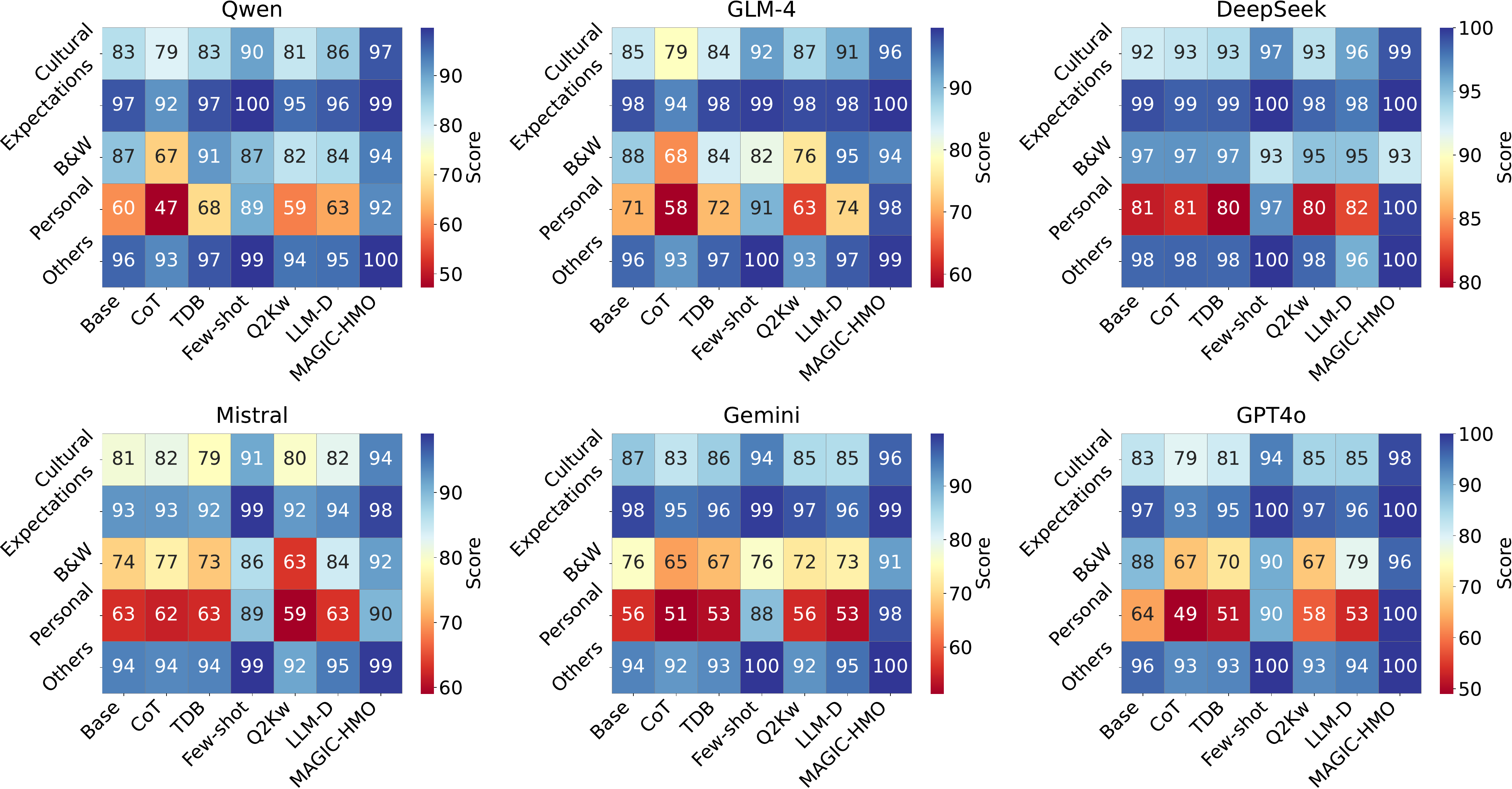}
    \caption{Fine-grained explicit objectives scores}
    \label{fig_comp_exp}
\end{subfigure}
\hfil
\begin{subfigure}{0.72\linewidth}
    \centering
    \includegraphics[width=\linewidth]{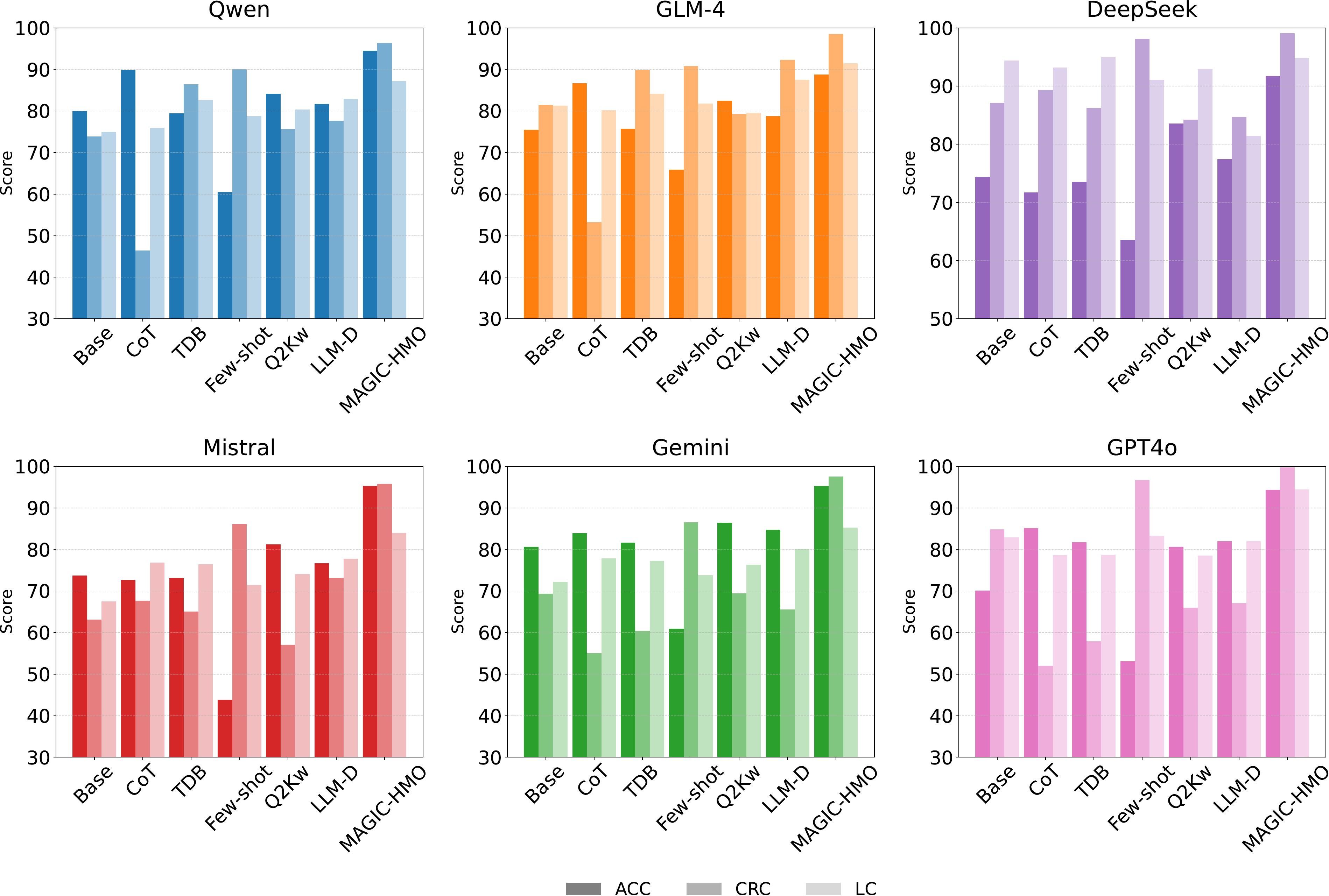}
    \caption{Fine-grained implicit objectives scores}
    \label{fig_comp_imp}
\end{subfigure}
\caption{Fine-grained comparison of explicit and implicit objective scores. (a) shows explicit scores (blue = high, red = low) across five aspects. (b) shows implicit results, where different colors indicate backbone models and shade depth reflects different metrics.}
\Description{A composite figure with two parts comparing explicit and implicit objective score across six backbone models (Qwen, GLM-4, DeepSeek, Mistral, Gemini, GPT-4o). (a) The top panel contains six heatmaps, one per model. Each heatmap shows scores for seven methods (Base, CoT, TDB, Few-shot, Q2Kw, LLM-D, MAGIC-HMO) across five aspects: cultural significance, parental expectations, BaZi & Wuxing, personal traits, and other requirements. Color ranges from red (low) to blue (high), with numerical values annotated in each cell. (b) The bottom panel contains six grouped bar charts, one per model. Each chart compares the same seven methods using three implicit metrics. Within each group, darker bars represent ACC, medium shades represent CRC, and lighter shades represent LC. Different base colors correspond to different backbone models, and higher bars indicate better performance.}
\label{fig_comp_exp_imp}
\end{figure}
To demonstrate the effectiveness of MAGIC-HMO, we compare it with base, reasoning-based (CoT~\cite{kojima2022large}, TDB~\cite{yang2024largelanguagemodelsoptimizers}), in-context (Few-shot~\cite{brown2020language}), retrieval-based (Q2Kw~\cite{jagerman2023query}), and traditional agent-based methods (LLM-D~\cite{lu2024llm}). Evaluation is conducted using both LLM-based and human metrics.
\subsubsection{LLM Evaluation Results}
As shown in Table~\ref{tab_com}, MAGIC-HMO consistently outperforms all baselines across different backbones, demonstrating its effectiveness in addressing the heterogeneous multi-objective optimization (HMO) problem while maintaining a better balance among objectives. We next analyze the performance of different methods in optimizing and balancing EUOs, IROs, and their combination. O1--O5 denote cultural significance, parental expectations, BaZi \& WuXing, personal traits, and other requirements, respectively.

Reasoning-based methods provide limited gains in EUOs' performance and even exhibit notable declines in some cases in Table~\ref{tab_com}. For example, CoT's EC (80.25) is lower than the Base’s EC (88.37) on GLM-4. A further analysis (Figure~\ref{fig_comp_exp}) reveals that reasoning-based methods perform well on common objectives (e.g., O1, O5) but struggle with others. This highlights the limitations of reasoning-based prompting in CNLG, as ~\cite{zhong2024let} discusses. 
The in-context methods, such as Few-shot, perform better than reasoning-based methods by introducing additional examples. For instance, Few-shot EC is 94.10, which is higher than 80.25 (CoT) on GLM-4.
However, other baselines still present notable limitations. Retrieval-based methods offer cultural knowledge that helps with objectives like O1 but introduce noise that negatively affects others (O2–O5). Consequently, Q2Kw often performs worse than Base in terms of EC and EC\_std (see Table~\ref{tab_com}). Traditional agent-based methods such as LLM-D struggle on O4 (Figure~\ref{fig_comp_exp}), indicating a need for more effective information expansion beyond user input. 
Most baselines perform poorly on complex and rarer objectives (e.g., O3, O4), highlighting their limited adaptability on diverse EUOs. 
In contrast, MAGIC-HMO demonstrates strong adaptability to diverse EUOs. It achieves the highest EC and the lowest EC\_std (Table~\ref{tab_com}), validating its effectiveness in optimizing multi-EUOs for CBN. This improvement is attributed to more fine-grained information expansion and examples that enhance query understanding and reduce noise. Additionally, explanation-oriented iterative dynamic optimization also provides more cues for verifying EUOs, which also improves the quality.

\begin{figure}[t!]
    \centering
    \includegraphics[width=0.8\linewidth,height=0.28\linewidth]{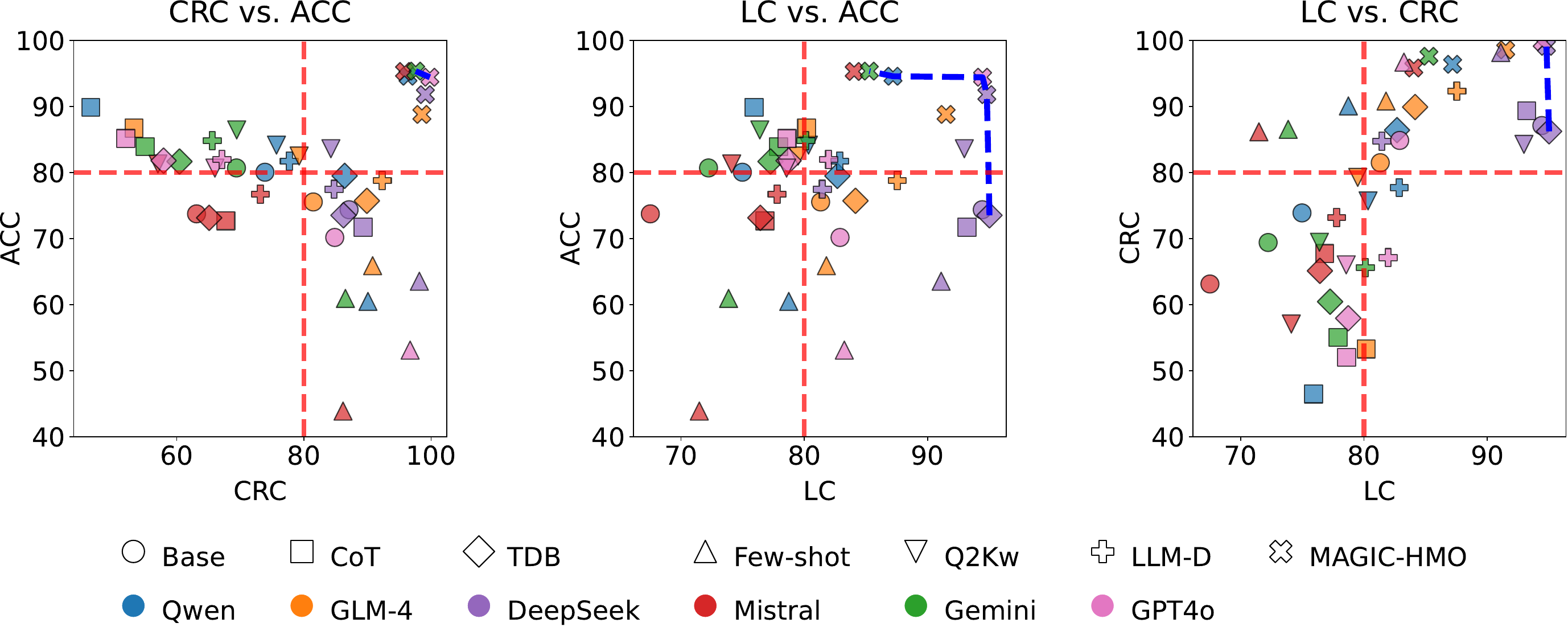}
   \caption{Pairwise comparisons of implicit metrics (ACC, CRC, LC). Blue line segments indicate the Pareto front, and dashed red lines mark the threshold of 80. Colors and marker shapes denote backbone models and methods, respectively.}
    \Description{A figure with three side-by-side scatter plots showing pairwise comparisons of implicit metrics (IRO). The three panels are: CRC vs. ACC (left), LC vs. ACC (middle), and LC vs. CRC (right). Each point represents a method--model combination, where marker shapes denote different methods (Base, CoT, TDB, Few-shot, Q2Kw, LLM-D, MAGIC-HMO) and colors denote backbone models (Qwen, GLM-4, DeepSeek, Mistral, Gemini, GPT-4o). Dashed red lines at the value of 80 on each axis indicate a reference threshold. Blue line segments connect the Pareto-optimal points in each plot, highlighting the best trade-off and balance between the two corresponding metrics.}
    \label{fig_comp_imp_pareto}
\end{figure}

As shown in Table~\ref{tab_com}, except for our framework, baselines perform unsatisfactorily at interpreting reliability (IC$<$90). 
Reasoning-based methods show higher IC\_std in Table~\ref{tab_com}. Figure~\ref{fig_comp_imp} shows that CoT exhibits unstable and poor ACC, CRC, and LC across backbones. Although TDB improves logic (↑LC), others remain unstable. 
Few-shot enhances explanation completeness (↑CRC) but increases hallucinations (↓ACC) due to noisy examples (e.g., Few-shot IC drops from 79.29 to 77.68 on GPT-4o, Table~\ref{tab_com}). 
Retrieval-based methods help mitigate hallucinations (↑ACC). Interesting, it also sacrifices completeness (↓CRC), but the reliability of explanation increases (e.g., Q2Kw IC from 76.29 to 80.05 on Qwen). This reflects that more relevant supplementary information can reduce input noise.
Traditional agent-based methods (e.g., LLM-D) improve logic and accuracy (↑LC, ↑ACC) of explanation but show inconsistent completeness (CRC), resulting in partial gains without fully optimizing IROs. 
In contrast, our explanation-oriented optimization strategy achieves superior results that cluster near the Pareto front with high scores (80+) (Figure~\ref{fig_comp_imp_pareto}), reflecting its effectiveness in optimizing multi-IROs.

Backbone differences impact performance partially (Table~\ref{tab_com}), including Chinese-oriented models generally outperform English ones, and models with stronger contextual understanding (e.g., DeepSeek, GPT-4o) perform better. For instance, under the Base, CC reaches 80.66 with Qwen but drops to 75.11 with Mistral. Most methods are affected by different backbones. Nevertheless, MAGIC-HMO always shows consistently strong performance, indicating its superior robustness. Although all methods boost diversity (DIV rise compared to Base in Table~\ref{tab_com}), MAGIC-HMO achieves the highest creative diversity, likely owing to more effective iterative guidance. Overall, MAGIC-HMO robustly optimizes EUOs, IROs, and their combination, validating its powerful capabilities in addressing the HMO problem.
\subsubsection{Human Evaluation Results}
\begin{figure}[t!]
    \centering
      \begin{subfigure}{0.42\linewidth}
        \centering
        \includegraphics[width=0.78\linewidth]{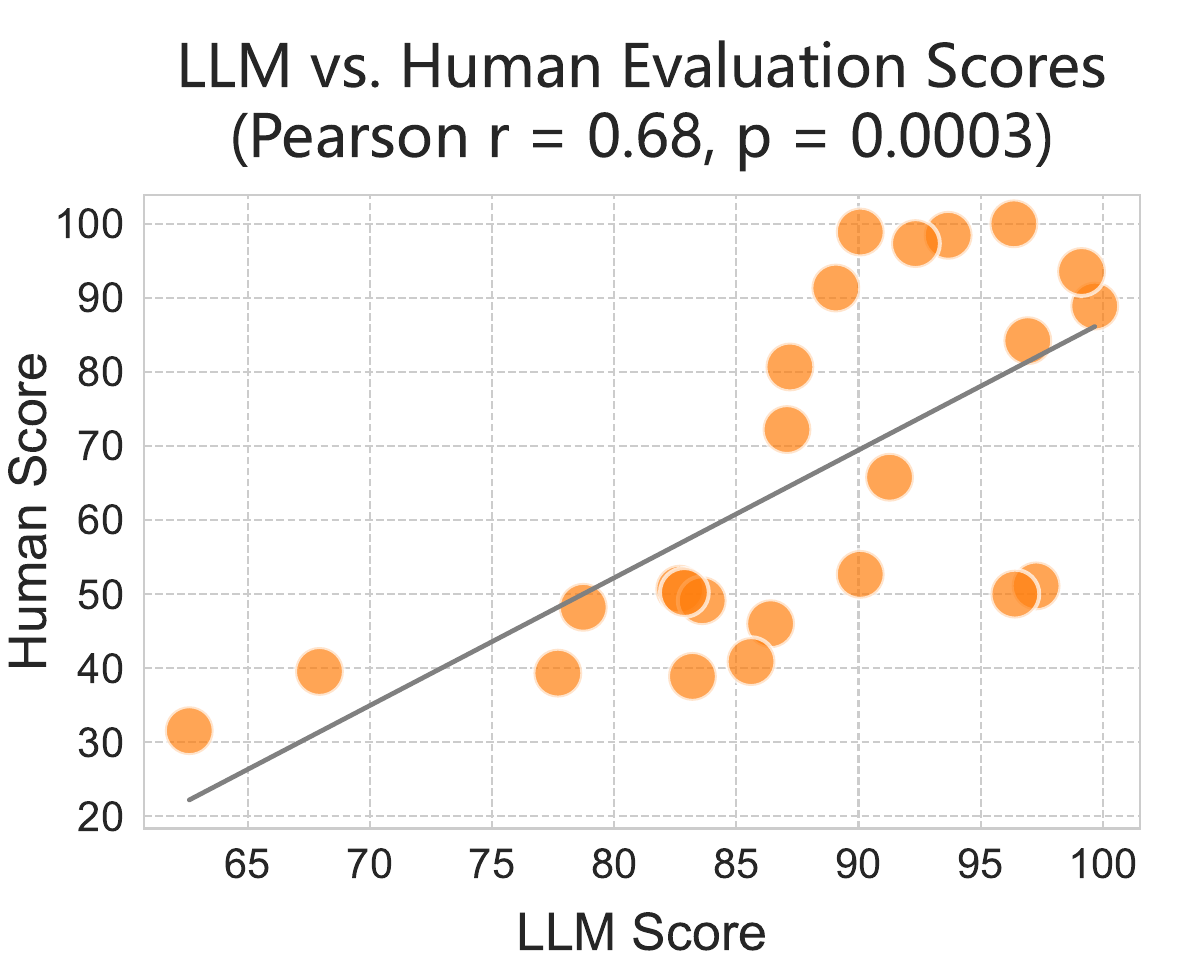}
        \caption{Correlation between LLM and human evaluation}
        \label{fig_human_llm}
    \end{subfigure}
    \hfil
    \begin{subfigure}{0.45\linewidth}
        \centering
        \includegraphics[width=0.73\linewidth]{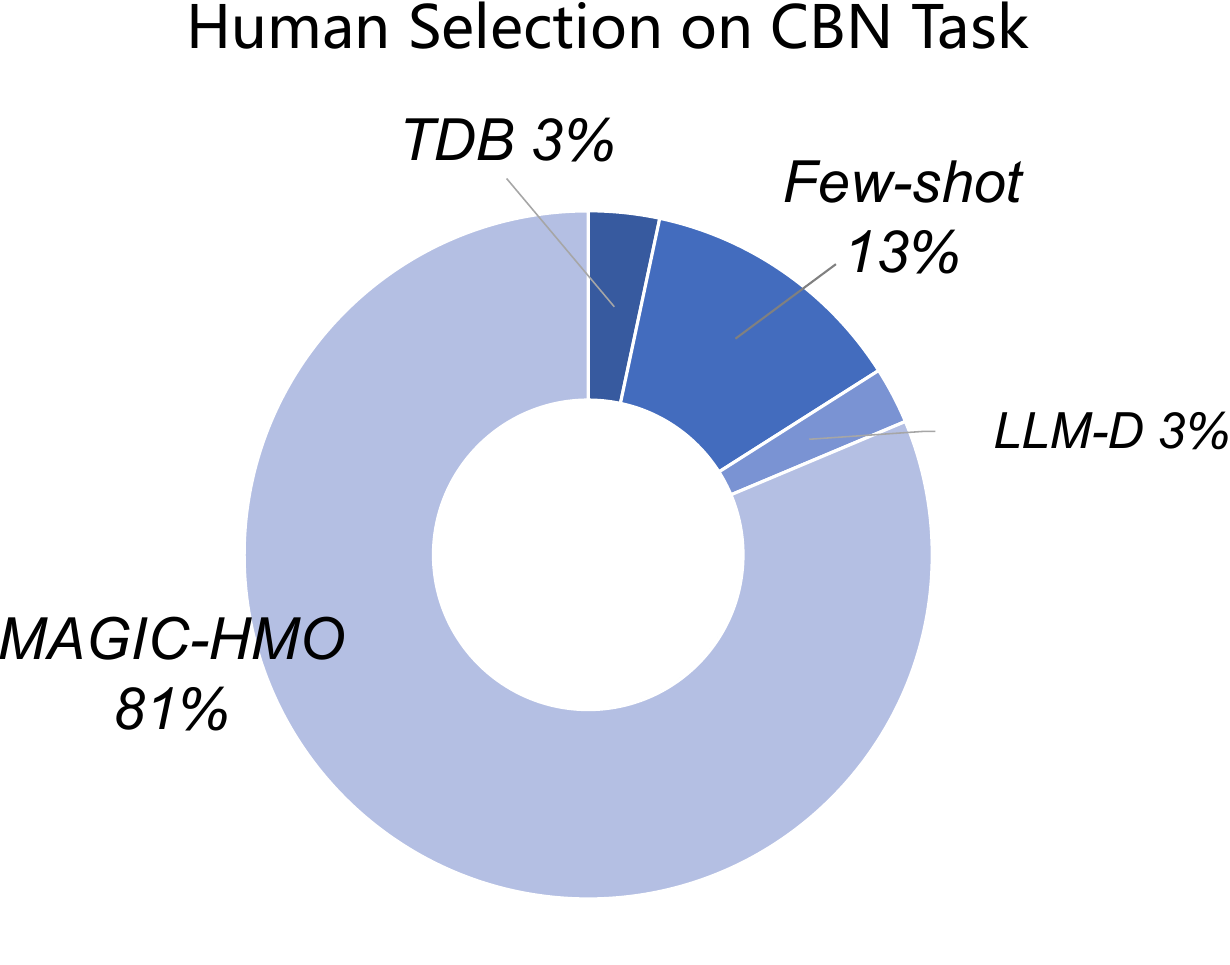}
        \caption{Human preference distribution}
        \label{fig_human_select}
    \end{subfigure}
    \caption{Human evaluation of TDB, Few-shot, LLM-D, and MAGIC-HMO on Qwen. (a) Correlation between LLM and human scores (Pearson r = 0.68, p = 0.0003). (b) Human preference distribution across methods.}
    \Description{A two-panel figure showing human evaluation results. (a) A scatter plot comparing LLM evaluation scores (x-axis) with human evaluation scores (y-axis). Each point represents a sample, and a fitted regression line is shown. The plot indicates a positive correlation between the two scores, with a Pearson correlation coefficient r = 0.68 and a p-value = 0.0003. (b) A donut chart showing the distribution of human preferences among four methods: TDB, Few-shot, LLM-D, and MAGIC-HMO. Each segment represents the proportion of times a method is selected as the best by human annotators, with MAGIC-HMO occupying the largest portion.}
    \label{fig_human}
\end{figure}

We compare our framework with three representative baselines (TDB, Few-shot, LLM-D) in human evaluation. Pearson correlation of $r=0.68(>0.5),p=0.0003(<0.001)$ (Fig.\ref{fig_human_llm}) indicates a statistically significant positive correlation between human and LLM evaluations. In Fig.~\ref{fig_human_select}, MAGIC-HMO is selected in 81\% of cases, substantially higher than Few-shot (13\%), LLM-D (3\%), and TDB (3\%). These results are consistent with the LLM-based evaluations, suggesting the effectiveness of our method and the reliability of the automatic metrics.
\subsubsection{Human Evaluation on Chinese Slogan Generation}
\begin{figure}[tpb!]
    \centering
    \includegraphics[width=0.5\linewidth]{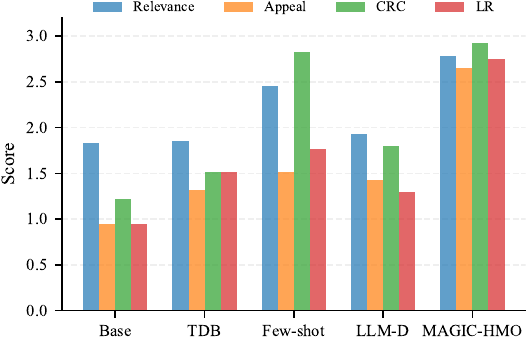}
    \caption{Human evaluation results on Chinese slogan generation.}
    \Description{A grouped bar chart showing the human evaluation results of five methods (Base, TDB, Few-shot, LLM-D, and MAGIC-HMO) across four metrics: relevance, appeal, constraint satisfaction (CRC), and linguistic readability (LR). Each method is represented by four bars corresponding to the four metrics, and the y-axis indicates the evaluation scores.}
    \label{fig_slogan_human}
\end{figure}

To further evaluate the generalizability of MAGIC-HMO beyond the CBN task, we conduct a human evaluation on a Chinese slogan generation task. Compared with CBN, slogan generation requires stronger alignment with user intent and more persuasive expression, serving as a complementary benchmark for Chinese short-form CNLG. Following the evaluation protocol used in CBN, we construct 20 slogan generation queries covering diverse user intents and scenarios. Three annotators are invited to evaluate the generated outputs on a 0--3 scale. To simplify the evaluation, we focus on three key aspects: \textbf{relevance} (alignment with user requirements), \textbf{appeal}, and \textbf{explanation quality}, where the latter is further measured by constraint satisfaction (CRC) and linguistic readability (LR). Figure~\ref{fig_slogan_human} reports the average scores across all samples and annotators for each method on the four evaluation metrics. MAGIC-HMO consistently achieves the best performance across all metrics, demonstrating clear advantages in both generation quality and explanation reliability. In particular, it yields substantial improvements in relevance and appeal, indicating better alignment with user intent and stronger persuasive capability. Meanwhile, its gains in CRC and LR suggest more complete and logically coherent explanations. In comparison, Few-shot performs relatively well on CRC but exhibits weaker overall balance. TDB and LLM-D show moderate improvements over the Base method, yet remain limited in jointly optimizing multiple objectives. Overall, these results further validate the effectiveness and generalizability of MAGIC-HMO beyond the CBN task.
\subsection{Ablation Study}
Ablation studies are conducted as follows. 
\begin{itemize}
\item \textbf{wo/~Imp\&Exp} removes implicit and explicit objectives' descriptions with their evaluation. 
\item \textbf{wo/~Imp} removes the implicit objectives' description and evaluation.
\item \textbf{wo/~Exp} removes the explicit objectives' description and evaluation. 
\item \textbf{wo/~evalGen} removes implicit and explicit evaluation. 
\item \textbf{wo/~R} removes the retrieval process. 
\item \textbf{wo/~evalR} removes the evaluation in retrieval.
\end{itemize}
\begin{table}[hb!]
\centering
\caption{Ablation Study Results with Qwen as the Backbone.}
\label{tab_abl}
\resizebox{0.8\linewidth}{!}{
\begin{tabular}{lccccccccc|cc}
\toprule
\multicolumn{1}{l}{\multirow{2}{*}{\textbf{Method}}}
& \multicolumn{3}{c}{\textbf{Explicit}}
& \multicolumn{6}{c}{\textbf{Implicit}}
& \multicolumn{2}{|c}{\textbf{Comprehensive}} \\
\cmidrule(r){2-4} \cmidrule(r){5-9} \cmidrule(r){10-11}
& EC ↑  & EC(c) ↑ & EC\_std ↓ & ACC ↑ & ACC(p) ↑ & CRC ↑ & LC ↑ & IC ↑ & IC\_std ↓ & CC ↑ & CC\_std ↓\\
\midrule
\textbf{MAGIC-HMO}	& \textbf{96.72} & \textbf{96.93} & \textbf{5.95} & \textbf{94.55} & \textbf{97.60} & \textbf{96.36} & 87.19 & \textbf{92.70} & \textbf{8.28}  & \textbf{94.71} & \textbf{4.64} \\
wo/ ImpExp	& 94.48 & -  & 9.81 & 63.25 & -  & 92.47 & \textbf{88.68} & 81.47 & 17.19 & 87.97 & 9.69 \\
wo/ Imp	    & 95.38 & -  & 7.10 & 91.05 & -  & 92.13 & 85.43 & 89.54 & 9.17  & 92.46 & 6.03 \\
wo/ Exp	    & 95.69 & -  & 7.57 & 92.50 & -  & 94.57 & 87.71 & 91.59 & 8.45  & 93.64 & 5.11 \\
wo/ evalGen	& 95.07 & -  & 7.51 & 89.70 & -  & 92.56 & 86.73 & 89.66 & 8.62  & 92.37 & 5.98 \\
wo/ R       & -  & 96.60 & - & - & 80.40 & - & - & - & - & - & - \\
wo/ evalR   & -  & 96.33 & - & - & 93.00 & - & - & - & - & - & - \\
\bottomrule
\end{tabular}}
\end{table}
\begin{figure}[hb!]
    \centering
      \begin{subfigure}{0.45\linewidth}
        \centering
        \includegraphics[width=\linewidth]{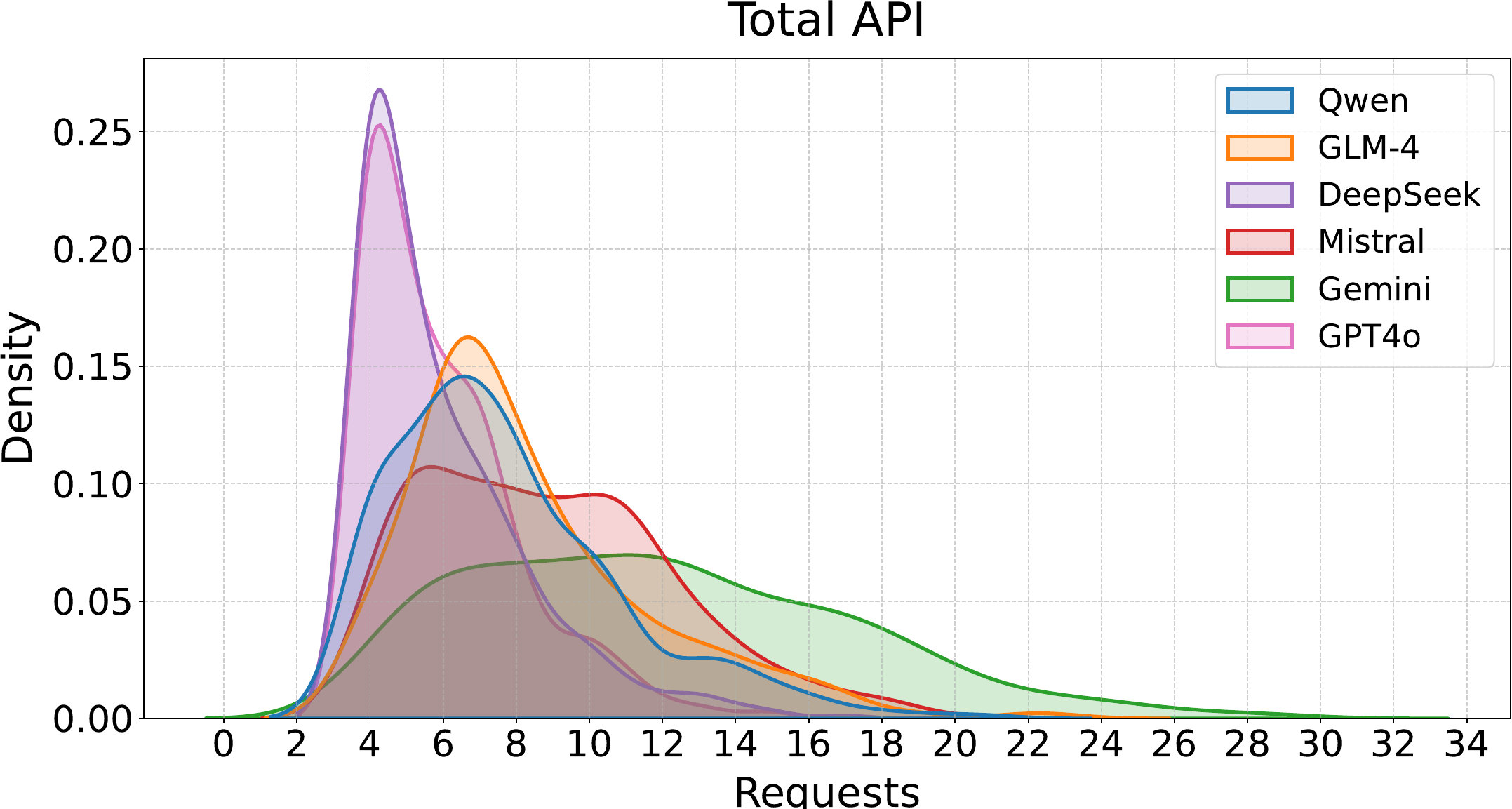}
        \caption{Total API requests}
        \label{fig_kde_overall}
    \end{subfigure}
    \hfil
    \begin{subfigure}{0.45\linewidth}
        \centering
        \includegraphics[width=\linewidth]{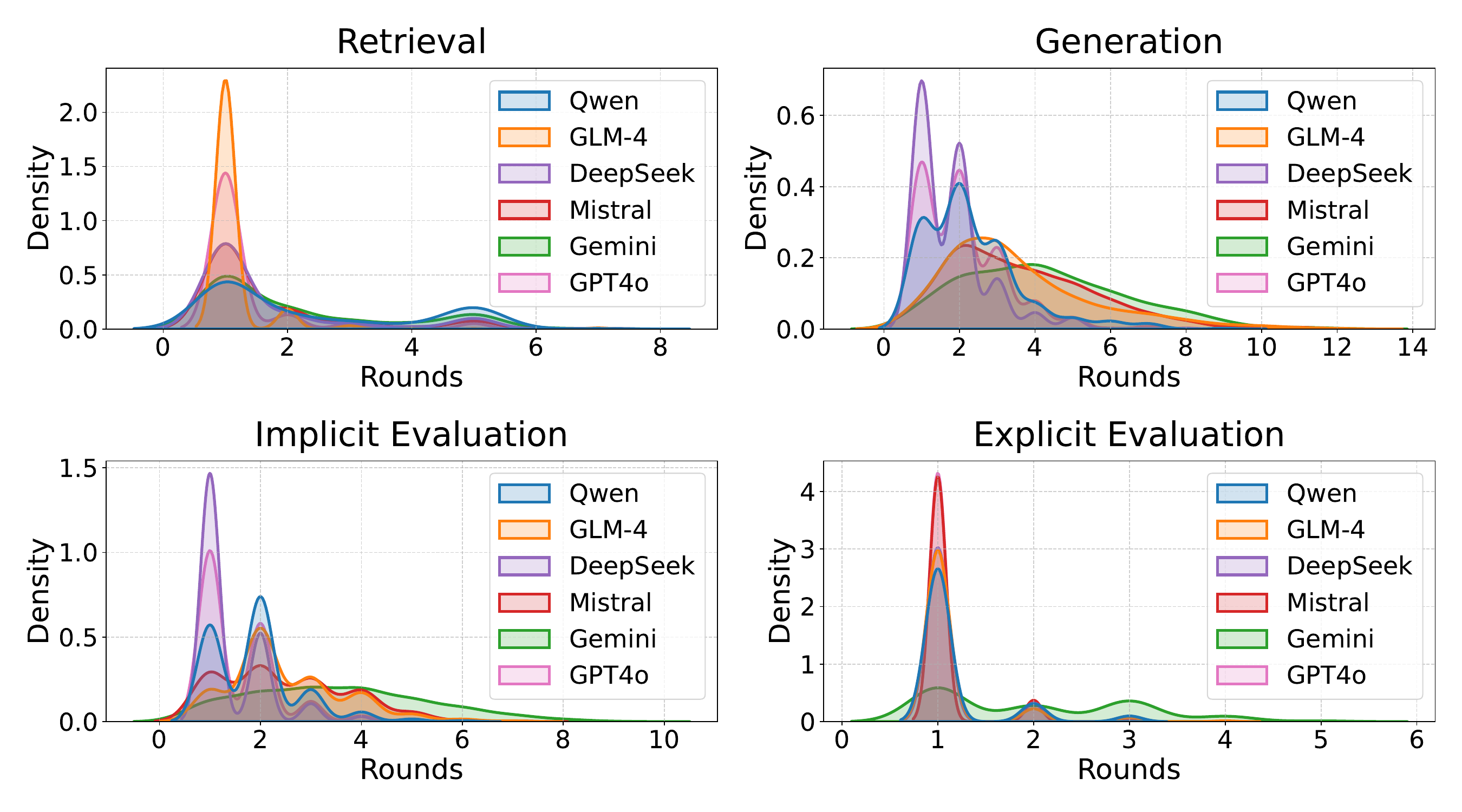}
        \caption{Iteration rounds by stage}
        \label{fig_kde_detail}
    \end{subfigure}
    \caption{Kernel density estimation (KDE) of interaction distributions across different LLM backbones with MAGIC-HMO. (a) shows API request counts over the full process; (b) shows iteration rounds in different stages, including Retrieval, Generation, and Implicit/Explicit Evaluation. Colors denote backbone models, and the y-axis represents estimated density.}
    \Description{A two-part figure showing kernel density estimation (KDE) plots of interaction distributions across multiple LLM backbones (Qwen, GLM-4, DeepSeek, Mistral, Gemini, GPT-4o). (a) A KDE plot of total API request counts, where the x-axis represents the number of requests and the y-axis represents estimated density. Each colored curve corresponds to a different backbone model. (b) Four smaller KDE plots showing the distribution of iteration rounds for different stages: Retrieval, Generation, Implicit Evaluation, and Explicit Evaluation. The x-axis represents the number of rounds, and the y-axis represents estimated density. Colors correspond to backbone models. Higher density values indicate regions with a greater concentration of samples.}
    \label{fig_kde}
\end{figure}
\begin{figure}[hb!]
    \centering
    \includegraphics[width=0.75\linewidth,height=0.22\linewidth]{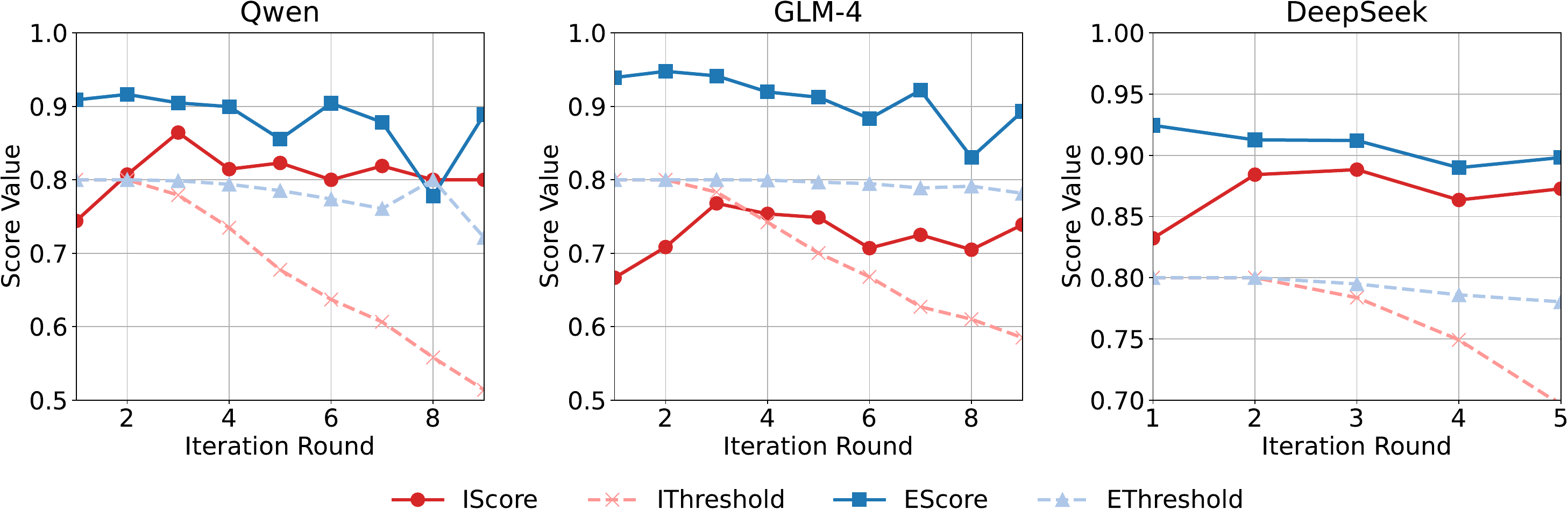}
    \caption{Iteration dynamics of implicit and explicit scores in MAGIC-HMO across backbone models. Red/blue lines denote implicit/explicit scores; solid/dashed lines indicate scores and thresholds.}
    \Description{A figure with three line charts showing the iteration dynamics of implicit and explicit scores for different backbone models (Qwen, GLM-4, DeepSeek). Each subplot corresponds to one model, with the x-axis representing iteration rounds and the y-axis representing score values. Red lines denote implicit scores and blue lines denote explicit scores. Solid lines represent the scores, while dashed lines represent the corresponding threshold values.}
    \label{fig_intera_line}
\end{figure}
The ablation results are summarized in Table~\ref{tab_abl}. In wo/~Imp\&Exp, most metrics exhibit a notable performance drop, highlighting the importance of detailed objective guidance and evaluation feedback in HMO. In particular, ACC drops sharply (from 94.55 to 63.25), demonstrating MAGIC-HMO’s effectiveness in mitigating hallucinations. Compared to MAGIC-HMO, the higher LC suggests that the additional information may introduce noise, which in turn slightly compromises logical rationality. A finer-grained analysis shows that wo/~Imp mainly affects implicit objective completion (IC from 92.7 to 89.54), while wo/~Exp more strongly impacts explicit objective balance (EC\_std from 5.95 to 7.57). This confirms the effectiveness of our approach in interpretation reliability and personalized constraints balance. Interestingly, wo/~Imp performs worse than wo/~Exp on EC, suggesting that optimizing IROs and EUOs is inherently interconnected, as lacking implicit objectives can also influence explicit objective fulfillment. MAGIC-HMO outperforms all ablation variants on CC and CC\_std, demonstrating its ability to achieve a better balance between these two types of objectives. 
Moreover, wo/~evalGen leads to a noticeable decline across metrics, indicating the effectiveness of our dynamic evaluation strategy. The wo/~evalGen and wo/~R perform poorly in ACC(p), revealing that the retrieval helps mitigate the LLMs' hallucination of domain knowledge. Additionally, wo/~evalR  degrades EC(c), indicating that evaluating retrieved content helps find the matcher knowledge for user needs. Overall, the results validate the effectiveness of different components of our multi-agent collaborative framework with dynamic iterative HMO.
\begin{figure}[ht!]
    \centering
     \begin{subfigure}{\linewidth}
        \includegraphics[width=\linewidth]{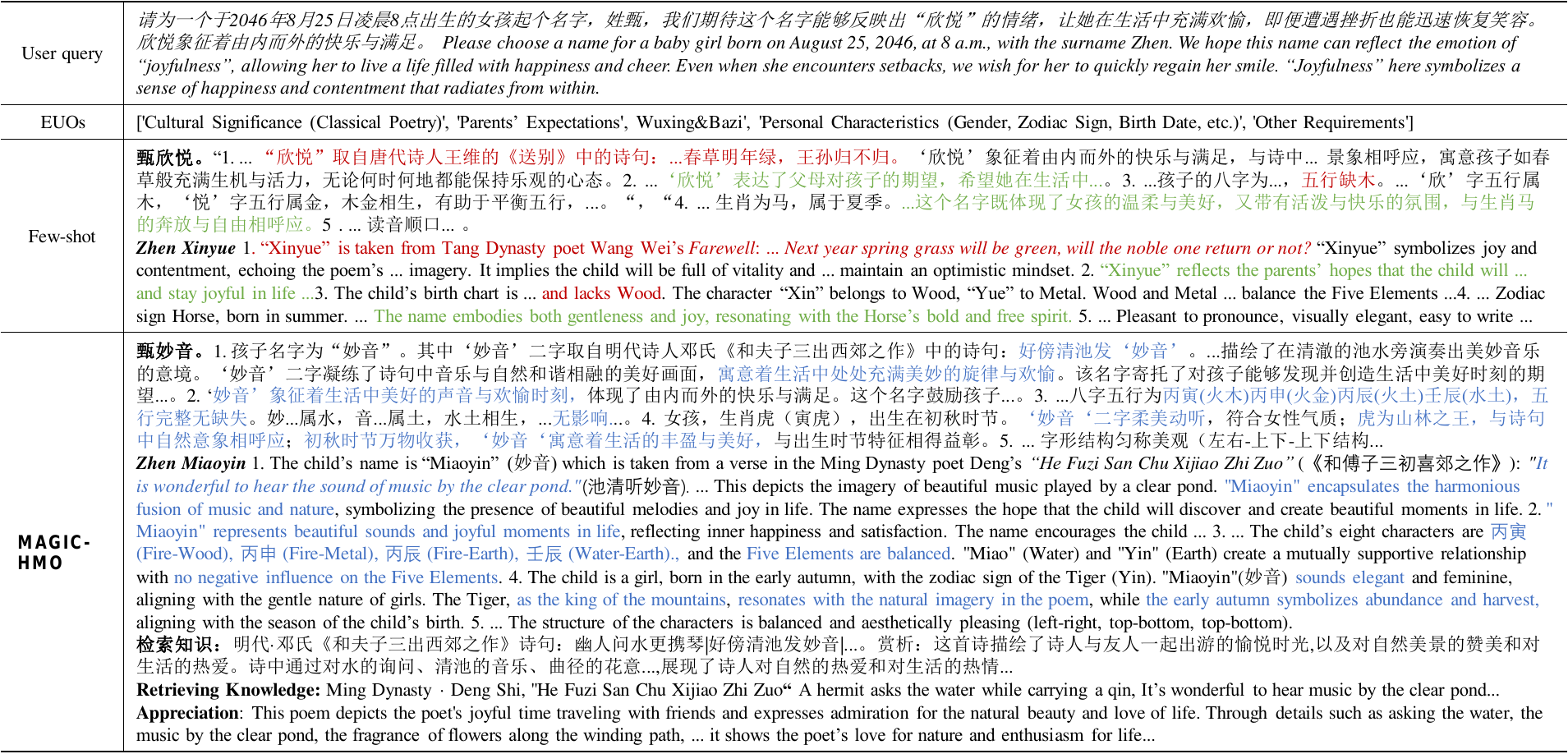}
        \caption{CBN task results}
        \label{fig_case1}
    \end{subfigure}
       \hfill
    \begin{subfigure}{\linewidth}
        \includegraphics[width=\linewidth]{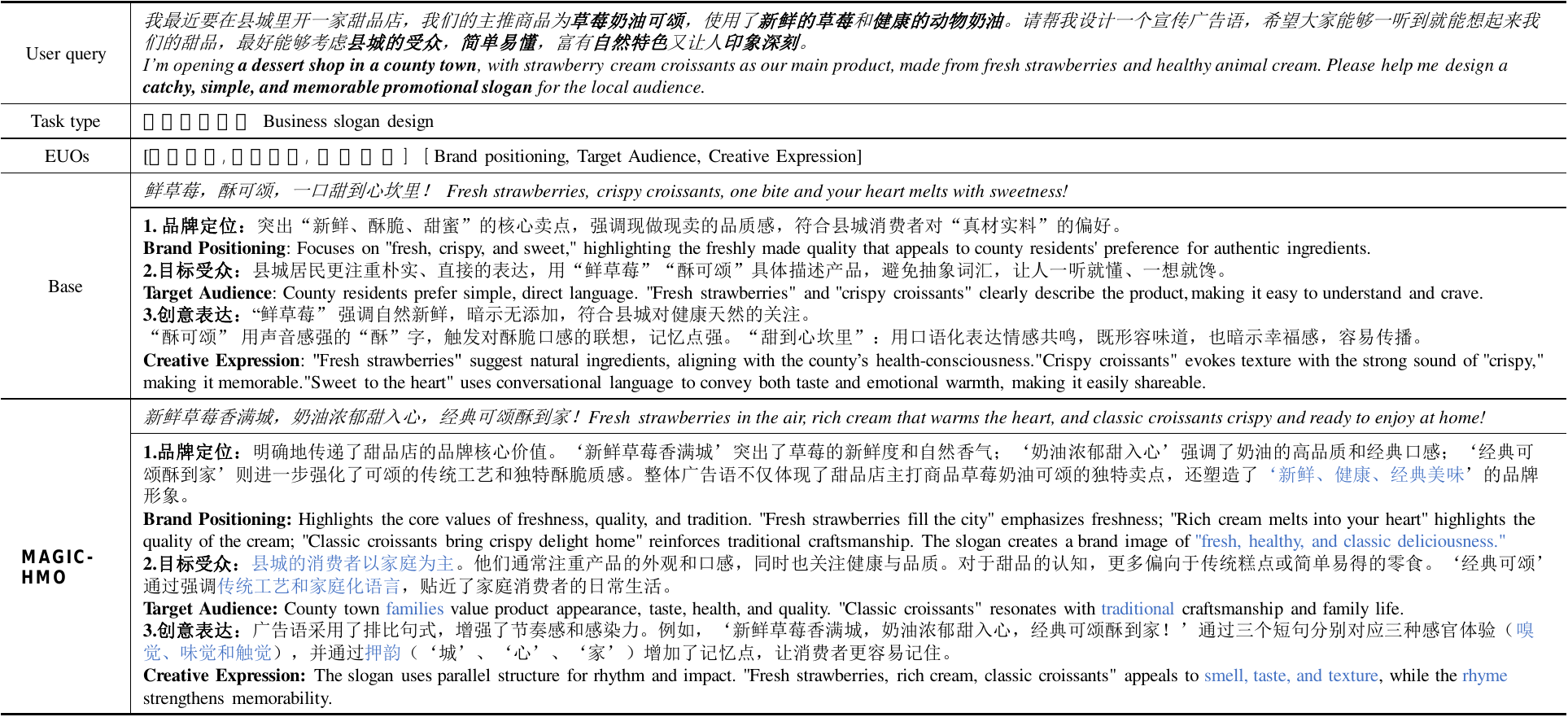}
        \caption{Slogan design task results}
        \label{fig_case2}
    \end{subfigure}
    \caption{Comparison of methods on the same query and backbone (DeepSeek). (a) CBN task results. (b) Slogan design task results. Red highlights factual or logical errors, green indicates vague explanations, and blue denotes content meeting EUOs with clear logic and explanations.}
    \Description{A two-part figure presenting case study comparisons of different methods on the same query using the DeepSeek backbone. (a) Results for a Chinese baby naming (CBN) task. The figure shows the user query, explicit user-specified objectives  (EUOs), and generated outputs from different methods, including Few-shot and NAMEGEn. The outputs are annotated with color highlights. (b) Results for a slogan design task, including the user query, task type, EUOs, and generated outputs from different methods. Across both panels, red highlights indicate factual or logical errors, green highlights indicate vague or insufficient explanations, and blue highlights denote content that satisfies explicit user objectives (EUOs) with clear logic and explanations.}
    \label{fig_cases}
\end{figure}
\subsection{Efficiency Analysis}
Figure~\ref{fig_kde} shows the interaction requests and density distribution at different stages of MAGIC-HMO. Figure~\ref{fig_kde_overall} exhibits the distribution of total API requests during the task. Most require only 5-10 requests, with a peak around 6, indicating high interaction efficiency and convergence. Figure~\ref{fig_kde_detail} further presents the interaction rounds at each stage. Retrieval interactions are mostly concentrated in early rounds, suggesting efficient retrieval. In the generation and evaluation stages, most samples require only 2–3 generation rounds and 1–2 evaluation rounds, reflecting efficient optimization. 
Overall, the distributions show clear concentration with a sparse tail, indicating that MAGIC-HMO completes most samples with limited interaction cost. Figure~\ref{fig_intera_line} illustrates the evolution of scores during the iterative process. The thresholds (dashed lines) gradually decrease over iterations, while the actual scores (solid lines) generally remain above the thresholds. This behavior suggests stable convergence and indicates that MAGIC-HMO can effectively support adaptive convergence control.
\subsection{Qualitative Evaluation}
We adopt DeepSeek for qualitative evaluation due to its strong performance in the CBN task. As illustrated in Figure~\ref{fig_case1}, MAGIC-HMO generates outputs that better align with user-specific objectives while providing more comprehensive and reliable explanations. In the CBN task, it effectively integrates diverse factors such as cultural references, parental expectations, and Bazi \& Wuxing. 
We further evaluate the generalization ability of MAGIC-HMO on Chinese slogan generation, as shown in Figure~\ref{fig_case2}. Even without explicitly specified EUOs, MAGIC-HMO can infer latent EUOs from the task context and user intent, producing outputs that better align with user needs. In this task, it captures key aspects such as brand positioning and target audience, highlighting its robustness in capturing user intents and its ability to handle diverse objectives across tasks.
\section{Conclusion}
This paper explores the unique challenges in Chinese short-form creative natural language generation (CNLG) and formalizes the problem as heterogeneous multi-objective optimization (HMO). We propose a training-free multi-agent framework, MAGIC-HMO, which employs an explanation-oriented strategy to address HMO. It jointly optimizes personalized user constraints and explanation reliability by dynamically verifying explanations and providing feedback. Experiments on the representative \emph{Chinese baby naming} task and an additional Chinese slogan generation task demonstrate that MAGIC-HMO consistently outperforms baselines across multiple LLM backbones. These results highlight the effectiveness and generalizability of MAGIC-HMO in producing higher-quality content that better satisfies user constraints with reliable explainability for Chinese short-form CNLG.
\bibliographystyle{ACM-Reference-Format}
\bibliography{references}
\end{document}